\documentclass{article}
\usepackage{cite}
\usepackage{amsmath,amssymb,amsfonts}
\usepackage{algorithm}
\usepackage{algpseudocode}
\usepackage{lipsum}

\usepackage{amssymb}
\usepackage{pifont}

\usepackage[flushleft]{threeparttable}

\usepackage{comment}

\usepackage{graphicx}
\usepackage{textcomp}
\usepackage{subcaption}
\usepackage{xcolor}
\def\BibTeX{{\rm B\kern-.05em{\sc i\kern-.025em b}\kern-.08em
    T\kern-.1667em\lower.7ex\hbox{E}\kern-.125emX}}
\usepackage[flushleft]{threeparttable}
\usepackage{arxiv}
\usepackage[english]{babel}
\usepackage{blindtext}
\usepackage{tabularx}
\usepackage[utf8]{inputenc} 
\usepackage[T1]{fontenc}    
\usepackage{hyperref}       
\usepackage{url}            
\usepackage{booktabs}       
\usepackage{amsfonts}       
\usepackage{nicefrac}       
\usepackage{microtype}      
\usepackage{lipsum}

\title{FedZip: A Compression Framework for Communication-Efficient Federated Learning}

\author{
  Amirhossein Malekijoo\\
  Department of Computer Science\\
  Semnan University\\
  Semnan, Iran \\
  \texttt{amirhossein.maleki1990@semnan.ac.ir} \\
   \And
Mohammad Javad Fadaeieslam \\
  Department of Computer Science\\
Semnan University\\
  Semnan, Iran \\
  \texttt{fadaei@semnan.ac.ir} \\
  
     \And
Hanieh Malekijou \\
The Iran University of Science and Technology\\
  Tehran, Iran \\
  \texttt{h\_malekijou@cmps2.iust.ac.ir} \\
  
     \And
Morteza Homayounfar \\
Amirkabir University of Technology\\
  Tehran, Iran \\
  \texttt{m.homayounfar@aut.ac.ir} \\
  
     \And
Farshid Alizadeh-Shabdiz \\
Boston University\\
  Boston, MA, USA \\
  \texttt{alizadeh@bu.edu} \\
     \And
\quad \quad \quad \quad Reza Rawassizadeh \\
\quad \quad \quad \quad Boston University\\
\quad \quad \quad \quad Boston, MA, USA\\
\quad \quad \quad \quad \texttt{rezar@bu.edu} 
}

\usepackage{cite}
\usepackage{amsmath,amssymb,amsfonts}
\usepackage{algorithm}
\usepackage{algpseudocode}
\usepackage{lipsum}

\usepackage{amssymb}
\usepackage{pifont}

\usepackage[flushleft]{threeparttable}

\usepackage{comment}

\usepackage{graphicx}
\usepackage{textcomp}
\usepackage{subcaption}
\usepackage{xcolor}
\begin{document}
\maketitle

\begin{abstract}
Federated Learning marks a turning point in the implementation of decentralized machine learning (especially deep learning) for wireless devices by protecting users’ privacy and safeguarding raw data from third-party access. It assigns the learning process independently to each client. First, clients locally train a machine learning model based on local data. Next, clients transfer local updates of model weights and biases (training data) to a server. Then, the server aggregates updates (received from clients) to create a global learning model. However, the continuous transfer between clients and the server increases communication costs and is inefficient from a resource utilization perspective due to the large number of parameters (weights and biases) used by deep learning models. The cost of communication becomes a greater concern when the number of contributing clients and communication rounds increases. 
In this work, we propose a novel framework, FedZip, that significantly decreases the size of updates while transferring weights from the deep learning model between clients and their servers. FedZip implements Top-z sparsification, uses quantization with clustering, and implements compression with three different encoding methods. FedZip outperforms state-of-the-art compression frameworks and reaches compression rates up to 1085$\times$, and preserves up to 99\% of bandwidth and 99\% of energy for clients during communication.
\end{abstract}


\section{Introduction}
Ubiquitous technologies such as smartphones and wearables have a tremendous influence on our lives. Complex approaches have been employed to allow these devices to provide cutting-edge performance of different tasks. 
Nowadays, many of their applications host machine learning algorithms. For example, smartphone applications use machine learning to sense audio \cite{Lane15a}, summarize daily user behavior \cite{Rawassizadeh16}, and recognize objects from images and videos~\cite{Mengwei18}. 
At the same time, deep neural networks (DNN) have revolutionized many disciplines in machine learning \cite{Sejnowski18}, and the demands placed on deep learning applications are continuously increasing \cite{Xu2019AFL}, even on battery-powered devices such as wearables~\cite{Lane15b}. This demand has led to the introduction of new hardware to support deep neural networks, usually by allocating an external chip or graphical processing unit (GPU) for this purpose, known as an AI chip~\cite{Rawassizadeh18b}. However, AI chips are expensive, and thus their inclusion in a wide variety of products is not cost-effective or affordable. In addition, as datasets’ sizes grow, machine learning models become more complex, with millions of parameters used in a simple DNN \cite{covidctnet}. Therefore, assigning a computation-intensive process to decentralized approaches became an inevitable alternative to centralized approaches. As a result, decentralized learning architectures, such as Federated Learning \cite{McMahan17a, Konecny15, McMahan17b}, on-device machine learning \cite{rawassizadeh2019b, keshavarz2020}, and edge computing \cite{Chen19} have been introduced to address this demand.

Federated Learning (FL) tries to address two important challenges \cite{McMahan17a}. First, it addresses tasks that DNN could perform on users’ devices, such as keyword spotting \cite{Leroy19} and image recognition \cite{Malekijoo19}. Second, since privacy is a major concern for end-users of mobile applications \cite{Rawassizadeh18b}, FL supports tighter privacy regulation~\cite{White12, Recital}. 

\begin{figure}[htbp]
\centering
\includegraphics[width=0.5\linewidth]{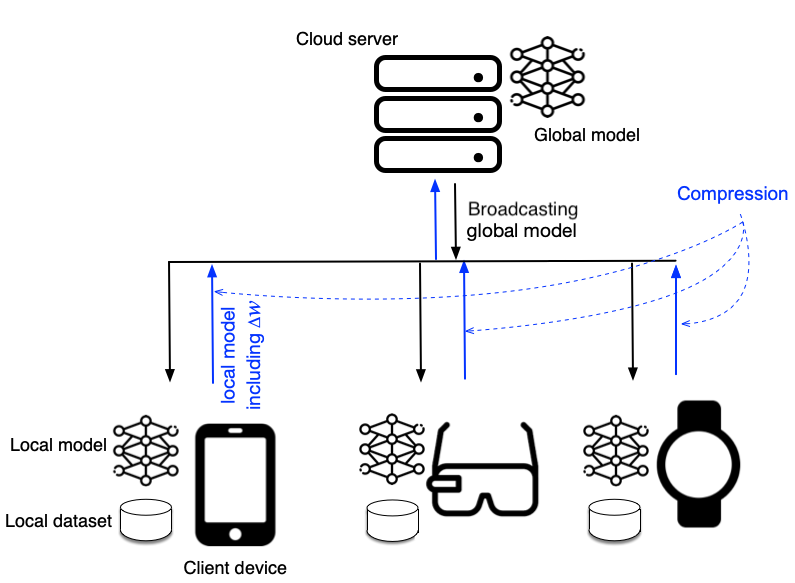}
\caption{The compression capability of the FedZip framework is highlighted relative to a traditional Federated Learning architecture. The $\Delta w$ builds the weight and bias updates that are exchanged between each client and the server.}
\label{fig:arch}
\end{figure}

In traditional systems, raw data is transferred from client devices to the server. Then, the server applies a DNN model to the data. The FL framework tries to reduce the amount of data involved by transferring only the model parameters and locally executed classifier results. Therefore, the raw data are not uploaded and classified in the central server. FL clients run a DNN model locally, employing the raw data stored locally, and builds a classifier model, an approach called \textit{local learning}. Then, the locally learned model is transferred to a server. On the server, a new global learning model is created (Figure \ref{fig:arch}) by aggregating clients’ updates, including model parameters (i.e. weights and biases, $\Delta w$, described in detail in next sections). The computational cost of training an accurate model arises from the raw data, which devolves the computational cost to client devices. To reduce this cost, the server builds a new global model and broadcasts it to all clients as the current model. This model is used as a base model for optimization in the next round of communication between the client and server. In other words, the weights (in the context of the neural networks) of clients’ models are aggregated inside the server and converged toward improving  accuracy in each round of communication.

The process of communication and aggregation in FL has two major drawbacks. First, the transmitted updates include all parameters in a DNN model imposes a heavy load of communication between clients and a central server. The upload rate from clients is a known network issue, measured by McMahan et al.~\cite{McMahan17a}
Second, the transmission of DNN models (i.e. the models themselves, without parameters) affects the battery of mobile or wearable devices, especially while uploading large models, such as R-CNN or U-Net~\cite{vuola2019mask}.
Settler et al.~\cite{Sattler18} suggested compression applied to downloads (downstream from the server) as well as upstream (upload) communications. Although this approach might reduce the communication cost, the convergence speed would also decrease significantly.

In this paper, we introduce a novel framework that integrates compression into FL architecture.The combination of three methods outperforms state-of-the-art algorithms in terms of accuracy and convergence speed.
Algorithms such as \cite{Bernstein18, Sattler18, Wen17} provide promising results, but either accuracy or convergence speed, or both are reduced for the sake of compression while reaching the highest compression rate. 

Our framework, FedZip, employs \emph{sparsification} based on Top-z pruning, then \emph{quantization} with k-mean clustering on model’s weights, and \emph{encoding} with Huffman Encoding and two other encoding methods, inspired by Huffman Encoding. As a result, FedZip reaches a higher compression rate besides lower degradation of accuracy than state-of-the-art algorithms \cite{Bernstein18, Sattler18, Wen17}. Blue arrows in Figure \ref{fig:arch} present which parts of the communication will be affected by FedZip. 

To implement sparsification, we employ a top-z sparsification method. To implement quantization, we optimally cluster weights and biases in update messages based on their distributions, which are going to transfer between each client and a server. To implement encoding, we use Huffman Encoding \cite{Huffman52}, a lossless compression algorithm. We also design two other encoding methods inspired by Huffman Encoding, that possess specific address tables to maximize the compression rate.

\section{Background and Related Work}
Our approach focuses on improving the performance of a distributed learning, by employing compression to improves the communication performance between client and server. Therefore, we discuss the literature on state-of-the-art methods in both fields. 
\subsection{Distributed Learning}
FL frameworks have been primarily studied on the convex objective function \cite{Konecny16a} with datasets that are evenly partitioned (balanced) in an independent and identically distributed (iid) fashion. However, real-world datasets are not necessarily balanced, and a machine learning model may not have a convex objective function. Therefore, an FL architecture should work for complicated DNN models, such as the Convolution Neural Network \cite{Krizhevsky12, Malekijoo19}. Besides, to be comprehensive, an FL architecture should consider three attributes for a dataset: non-iid, unbalanced, and non-convex \cite{McMahan17b}. The non-convex form of problems with a partitioned dataset in a non-iid fashion was proposed by McMahan et al.~\cite{McMahan17a} for benchmark datasets (e.g. MNIST~\cite{LeCun18}, CIFAR-10~\cite{Krizhevsky09}, and LSTM on a language dataset~\cite{W.Shakespeare}).

FL optimization methods focus on incremental optimization methods, ,mostly Gradient Descent. Chen et al. \cite{Chen16} have demonstrated a synchronous large-batch Stochastic Gradient Descent (SGD) works more efficiently than an asynchronous approach. Synchronization in this context means that all clients aggregate their updates at the same time. In the asynchronous approach, however, this process is extended over time to alleviate the loads. Bonawitz et al.~\cite{Bonawitz19} have introduced a framework based on the recommendations of Chen et al.~\cite{Chen16} to alleviate the loads with asynchronous approaches. There are two well-known methods, suggested by McMahan et al.~\cite{McMahan17a, McMahan17b} and Konecnye et al.~\cite{Konecny15}. FedSGD and FedAvg are the closest model to our work, thus in our experiments, we will study them in more detail.

The first method, FedSGD \cite{McMahan17a}, is an SGD-inspired approach. FedSGD employs a random subset of clients (instead of all participants) to create a global learning model for just one round. Each client then receives the global model in each round and performs an SGD-based optimization on their dataset, with all clients' data points used for optimization. FedSGD is computationally efficient, and its convergence speed increases in each round, but it does require a lot of rounds of training to produce an efficient model. Note, In this context, by round, we mean upload from the client to the server, global model averaging, and download from the server. It is different than an epoch, which focuses on the execution of the neural network algorithm (centralized model) and changing its weights and biases of the local model.

The second method, FedAvg \cite{McMahan17a}, provides a more generalized perspective than FedSGD. It allows each client to perform an SGD optimization in either more than one batch on local data or to perform more iterations, and instead of gradients, it exchanges the delta weights with the server. On the server, model averaging is performed to generate a global model. Authors report that if each client can perform a local SGD on their batched dataset with more than one training pass (to reduce the communication cost), the model’s accuracy does not change significantly. Their experiments demonstrate that FedAvg is robust to unbalanced and non-iid data distribution. The convergence of FedAvg on non-iid datasets has been studied by Li et al. \cite{Xiang19}, who propose using the decay technique for the global learning rate to reach optimal convergence. If all clients of the FedSGD model start from the same global model (i.e. the weights which initialize the clients in each round), averaging the gradients is equivalent to averaging the weights themselves. However, the intuition behind FedAvg, and its general perspective toward employing the average model, is that using averaged weights from the same global model does not harm the model's performance~\cite{McMahan17a}.

Several studies have examined the effects of client selection \cite{Nishio19}, upload cost reduction for clients \cite{Konecny16b}, and budget-constrained optimization \cite{Wang18a}. More recently, a dynamic averaging method based on local model divergence criterion has been suggested by Kamp et al.~\cite{Kamp18}, and Geyer et al.~\cite{Geyer18} have evaluated the FL algorithm's robustness to different privacy attacks.
Since our objective is to focus on efficient communication and preserving the accuracy, comparing FedZip with these approaches is out of the scope of this paper.


\begin{table}[htbp]
\begin{center}
  \caption{Comparison of state-of-the-art distributed learning methods with compression and FedZip.}
  \label{table:overview}

 \footnotesize

  \begin{tabular}{|c|c|c|c|c|}
  \hline
  \textbf{Methods} & \textbf{CR}  & \textbf{Robust to}  & \textbf{Compression} & \textbf{Auxiliary}\\ 
    &  & \textbf{non-iid} & \textbf{Route} & \textbf{Techniques }\\
    \hline
    \midrule
    \ SignSGD~\cite{Bernstein18} & 32 & \ding{55} & Bidirectional & \checkmark\\
    \hline
    \ TernGrad~\cite{Wen17} & \textless 32 &  \ding{55} & Bidirectional & \checkmark\\
    \hline
    \ QSGD~\cite{Alistarh17} & \textless 32  & \ding{55} & Upstream & \checkmark\\
    \hline
    \ ATOMO~\cite{Wang18a} & 32  & \ding{55} & Upstream  & \checkmark\\
    \hline
    \ GD~\cite{Aji17} &	\textgreater 32 & \checkmark & Upstream & \ding{55}\\
    \hline
    \ DGC~\cite{Lin18} & \textgreater 32 & \checkmark & Upstream & \checkmark\\
    \hline
    \ STC~\cite{Sattler18} & \textgreater 32 & \checkmark & Bidirectional & \checkmark\\
    \hline
    \ FedAvg~\cite{McMahan17a} & \textless 32\tnote{1} & \checkmark & Upstream\tnote{2} &\ding{55}\\
    \hline
    \ \textbf{FedZip} &	\textgreater 32 &	\checkmark & Upstream\tnote{3} & \ding{55} \\
    \hline
  \bottomrule
  \end{tabular}

  \begin{tablenotes}
    \item[1]1. With auxiliary techniques, FedAvg and FedZip reach a higher CR. Adding auxiliary techniques is out of the scope of this paper.
    \item[2]2. if auxiliary techniques, such as delay communication, are used FedAvg converts to a bidirectional framework.
    \item[3]3. if auxiliary techniques are used FedZip can be bidirectional (like FedAvg). 
  \end{tablenotes}
\end{center}
\end{table} 

\subsection{Compression}
The contributions mentioned to this point have focused on reducing the communication costs of FL architectures, but another body of work has focused on compression of the updates between the server and clients while maintaining the accuracy of a DNN model. These methods are referred to as ``Dense Quantization.'' 

Han et al. \cite{Han15} propose a dense quantization method that utilizes pruning, quantization, and Huffman Encoding to quantize the weights of a DNN with minimal degradation of accuracy. They report a compression rate of up to $49\times$ without a loss of accuracy; the model's training speed and energy efficiency were also improved, where increases up to $4.2\times$ and $7\times$, respectively. Wen et al.~\cite{Wen17} stochastically quantize gradients to ternary and introduce ``TernGrad'' . It achieves a $16\times$ compression rate in decentralized learning, though the accuracy decreased by more than $2\%$ in a complex DNN model.

Alistarh et al.~\cite{Alistarh17} propose QSGD, a method to investigate the relationship between the gradient precision and  accuracy of a distributed learning structure.QSGD provides a convergence guarantee by considering a threshold as a lower bound. Experiments demonstrate that the information-theoretic lower bound would violate, if the sparsification passes a lower bound threshold. QSGD emphasizes adding a variable variance that can be controlled. However, its representative value of quantization does not follow the weight distribution~\cite{Alistarh17}. 

Bernstein et al. \cite{Bernstein18} have introduced a method for compressing the communicative updates in non-convex problems. They propose a form of gradient compression, which takes the sign of each coordinate from the stochastic gradient vector and aggregates the gradient signs of the clients by majority voting. They assume the default variable in a normal DNN is float32, so every variable is quantized to $1$ bit (binary sign) $32\times$ compression. Their approach resulted in an accuracy of $2\%$ lower than SGD in the test phase. 
Siede et al. \cite{Seide14} introduce a dense quantization method that makes possible 1-bit quantization for weight updates. They parallelize SGD by carrying the error from each previous round (e.g. errors in the round one and two will be carried to round three) to ensure high convergence speed while still considering all gradients during model training.

Sattler et al. \cite{Sattler18} propose a scheme that includes multiple techniques, including communication delay, gradient sparsification, and binarization to reduce the number of elements that must be considered. The authors used sparsification method and encoding method to achieve a greater than $32\times$ compression rate. Abdi et al.~\cite{Abdi} investigate the effect of adding independent dither to improve the desirability of the quantization error statistical behavior. Their work, however, imposes a further calculation complexity on a large scale system. Wang et al.~\cite{Wang18b} have introduced ATOMO, which uses a singular value decomposition method (SVD) to sparsify a tensor of neural network gradients and mitigate the communication load between clients and a server. Aji et al. \cite{Aji17} propose a method that maps the $99\%$ smallest gradients (by absolute magnitude) to zero and then transmits the sparse matrices to a server. Their method is combined with quantization to further improve compression. Shaohuai et al. \cite{Shaohuai19} proposed a global top-$z$ sparsification model that suffers from accuracy degradation. DGC \cite{Lin18} creates an address table for the position of the non-zero values of the delta weights and communicates the address table and $1$ bit for two infrequent classes. However, because of using gradient weights without considering the distribution, and use of '$0$` as the representative of the most frequent class (instead of a real representative of the weights’ distribution), their approach requires auxiliary methods such as warm-up training, momentum correction and fine-tuning to reach an accuracy close to the baseline.

A quantitative comparison between  state-of-the-art compression and our approach will be analyzed later in our experiments.

\begin{figure*}[tb!]
  \centering
  \captionsetup[subfigure]{labelformat=empty}
  \subcaptionbox{{\fontfamily{phv}\selectfont
    \footnotesize{Layer1: Conv2D}}}[.24\linewidth][c]{%
    \includegraphics[width=\linewidth]{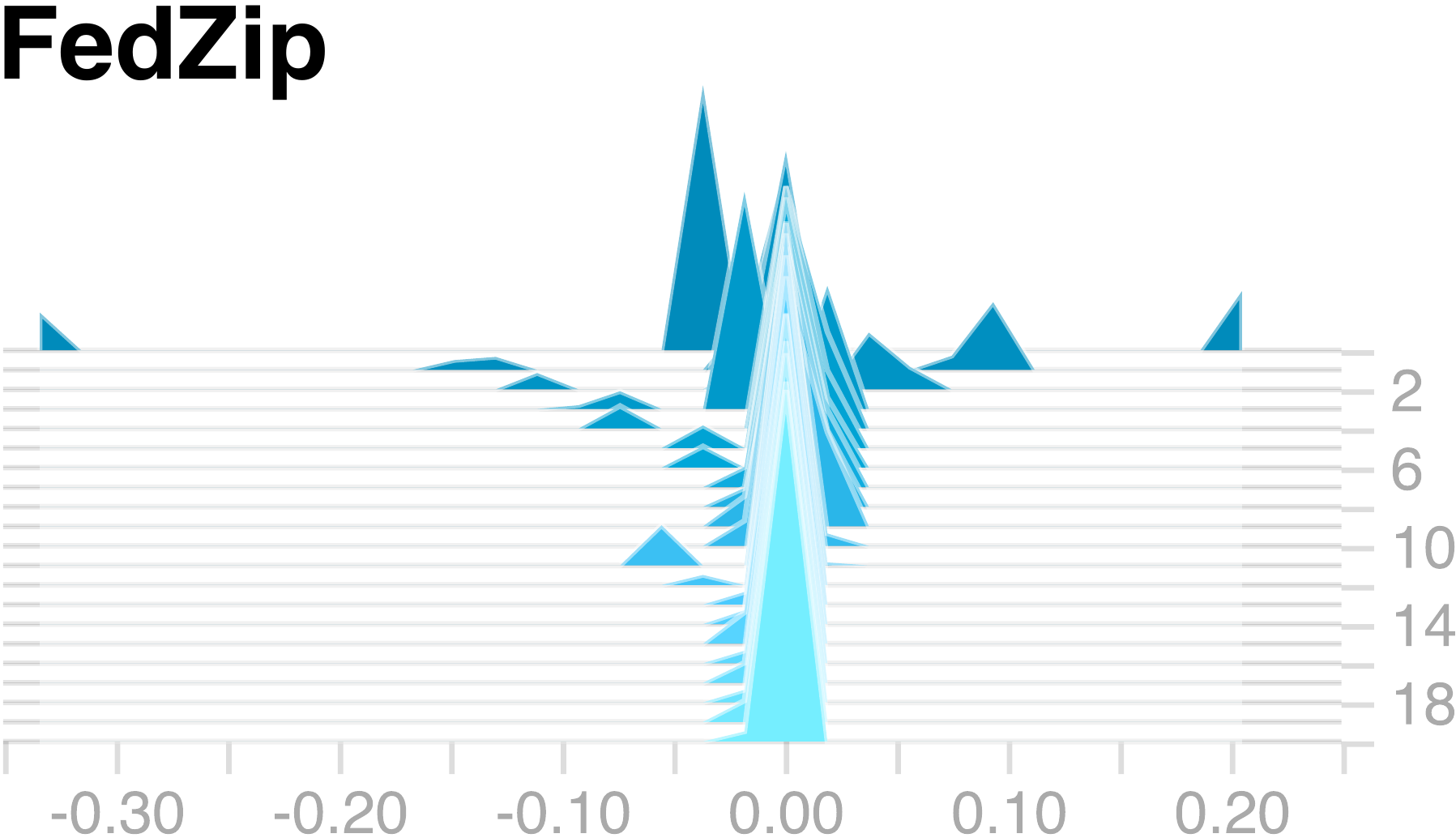}}
  \subcaptionbox{{\fontfamily{phv}\selectfont
    \footnotesize{Layer2: Conv2D}}}[.24\linewidth][c]{%
    \includegraphics[width=\linewidth]{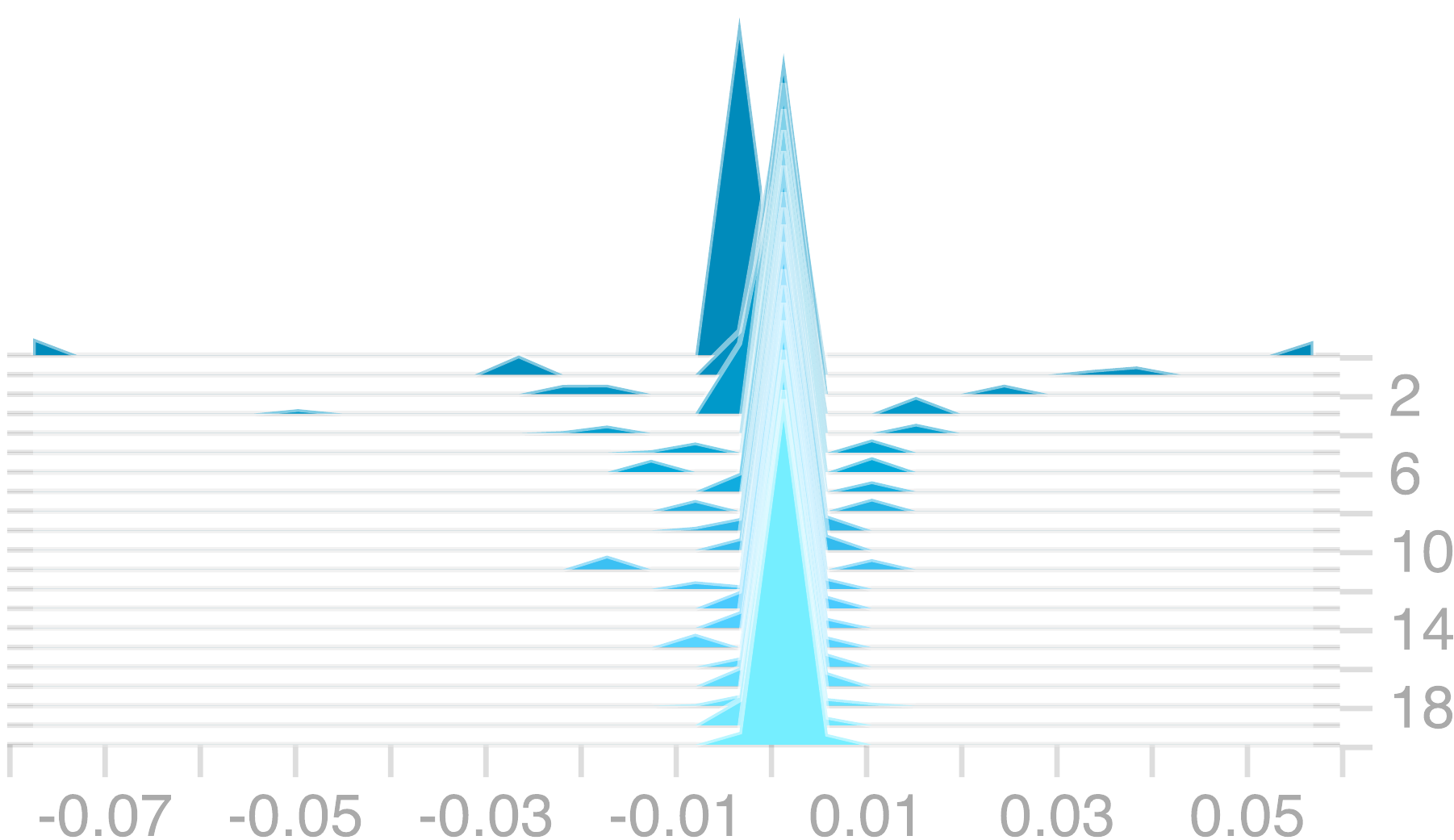}}
  \subcaptionbox{{\fontfamily{phv}\selectfont
    \footnotesize{Layer3: Dense}}}[.24\linewidth][c]{%
    \includegraphics[width=\linewidth]{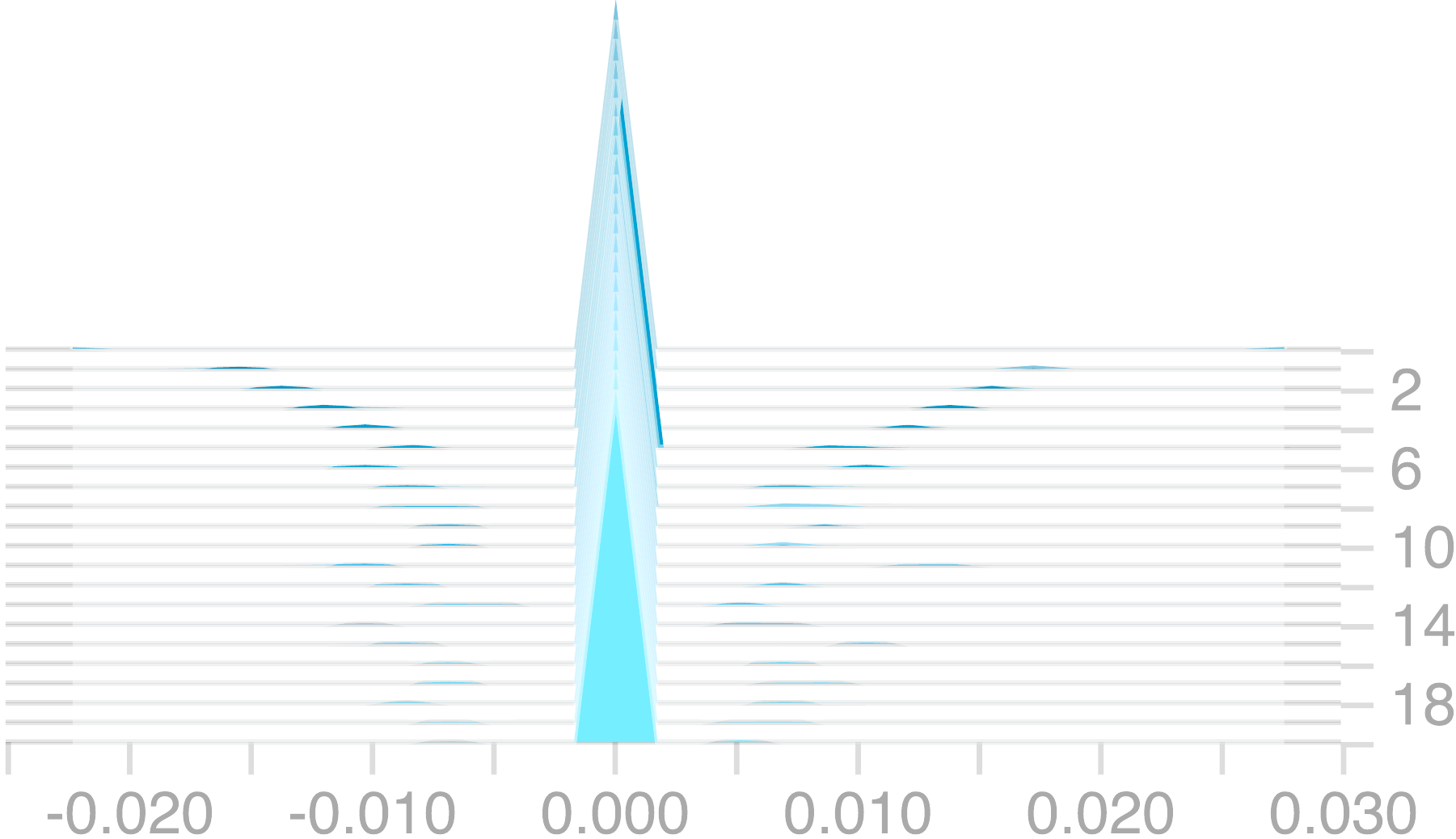}}
  \subcaptionbox{{\fontfamily{phv}\selectfont
    \footnotesize{Layer4: Dense}}}[.24\linewidth][c]{%
    \includegraphics[width=\linewidth]{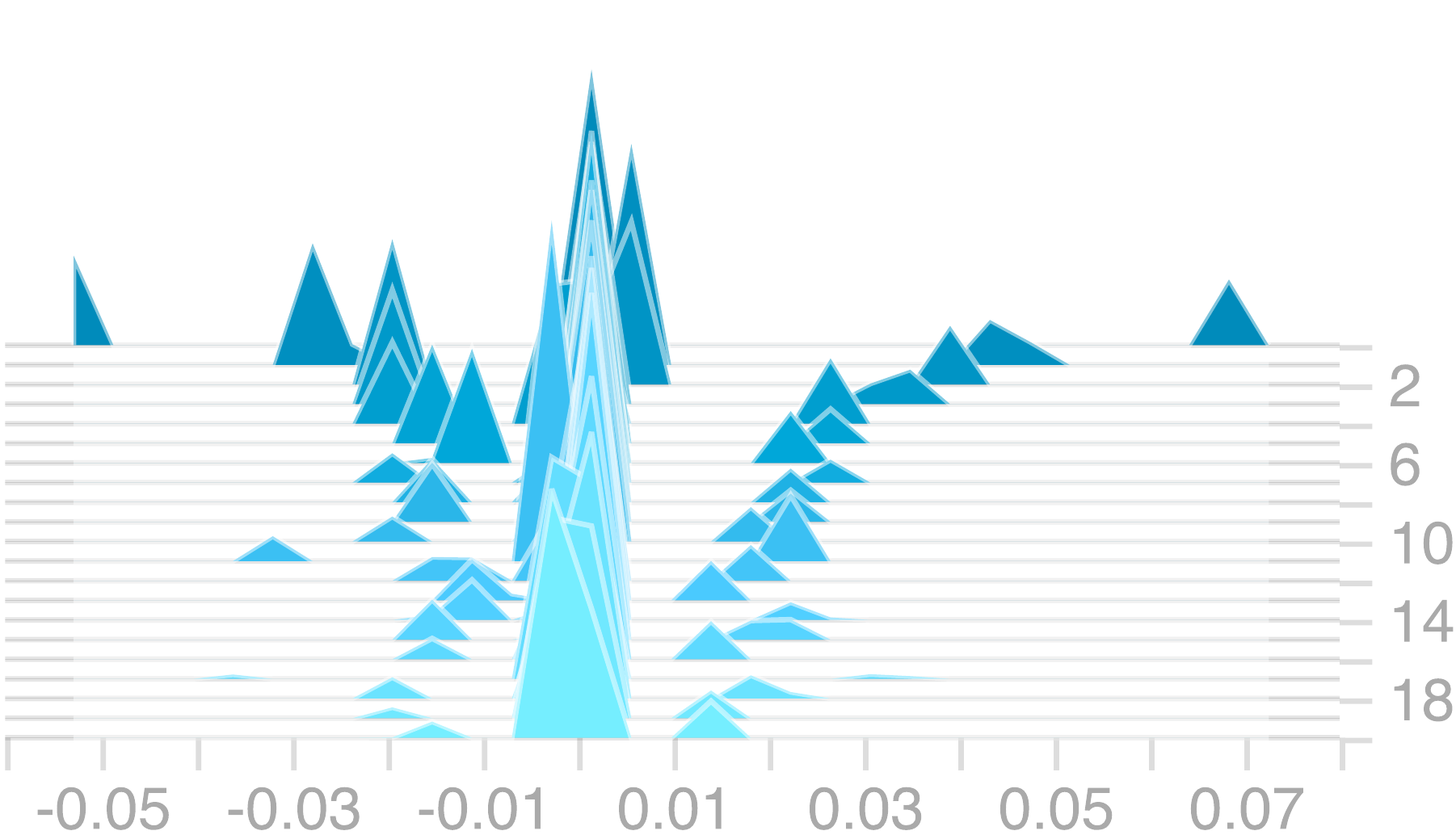}}

  \bigskip

  \subcaptionbox{{\fontfamily{phv}\selectfont
    \footnotesize{Layer1: Conv2D}}}[.24\linewidth][c]{%
    \includegraphics[width=\linewidth]{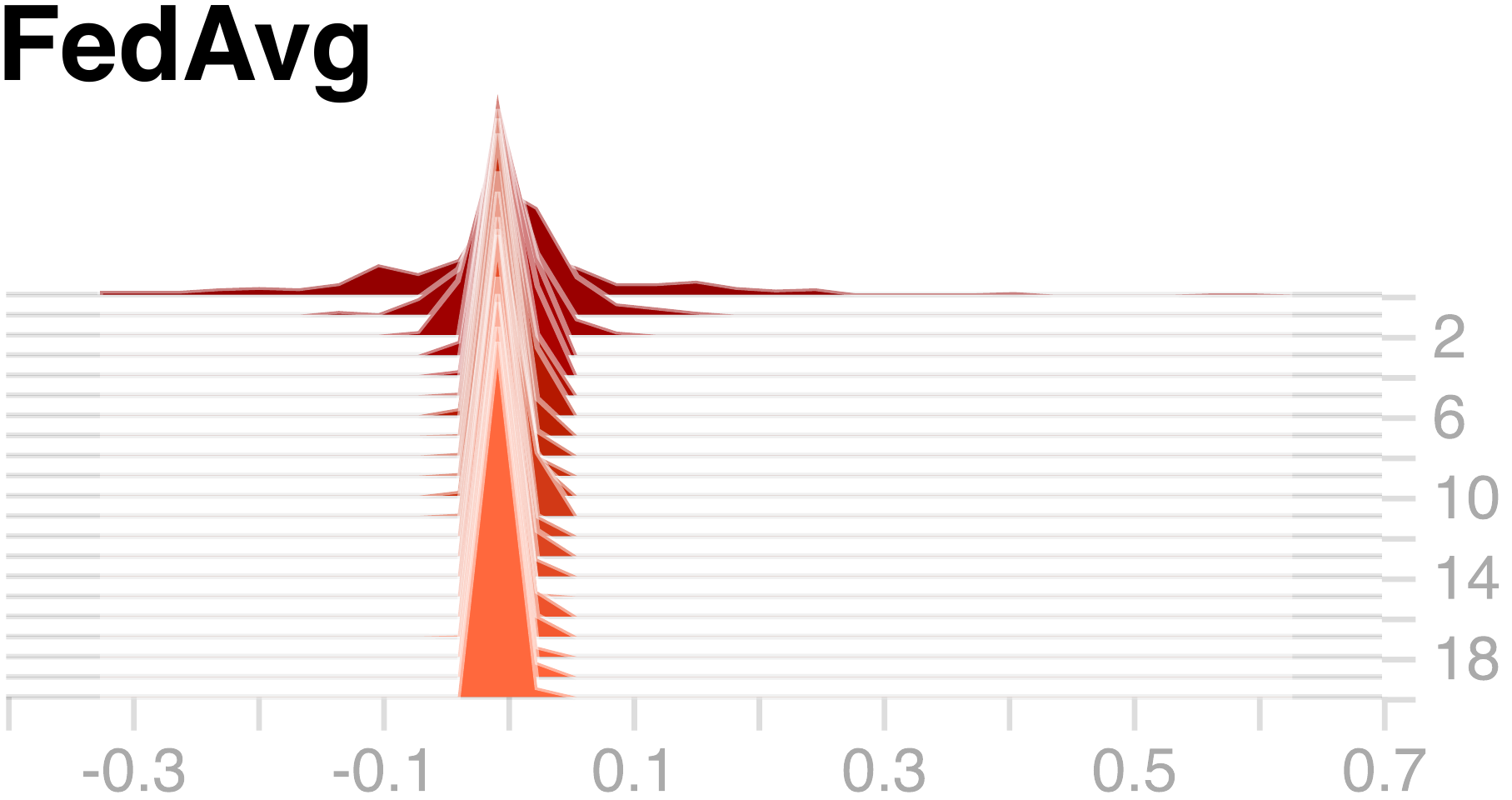}}
  \subcaptionbox{{\fontfamily{phv}\selectfont
    \footnotesize{Layer2: Conv2D}}}[.24\linewidth][c]{%
    \includegraphics[width=\linewidth]{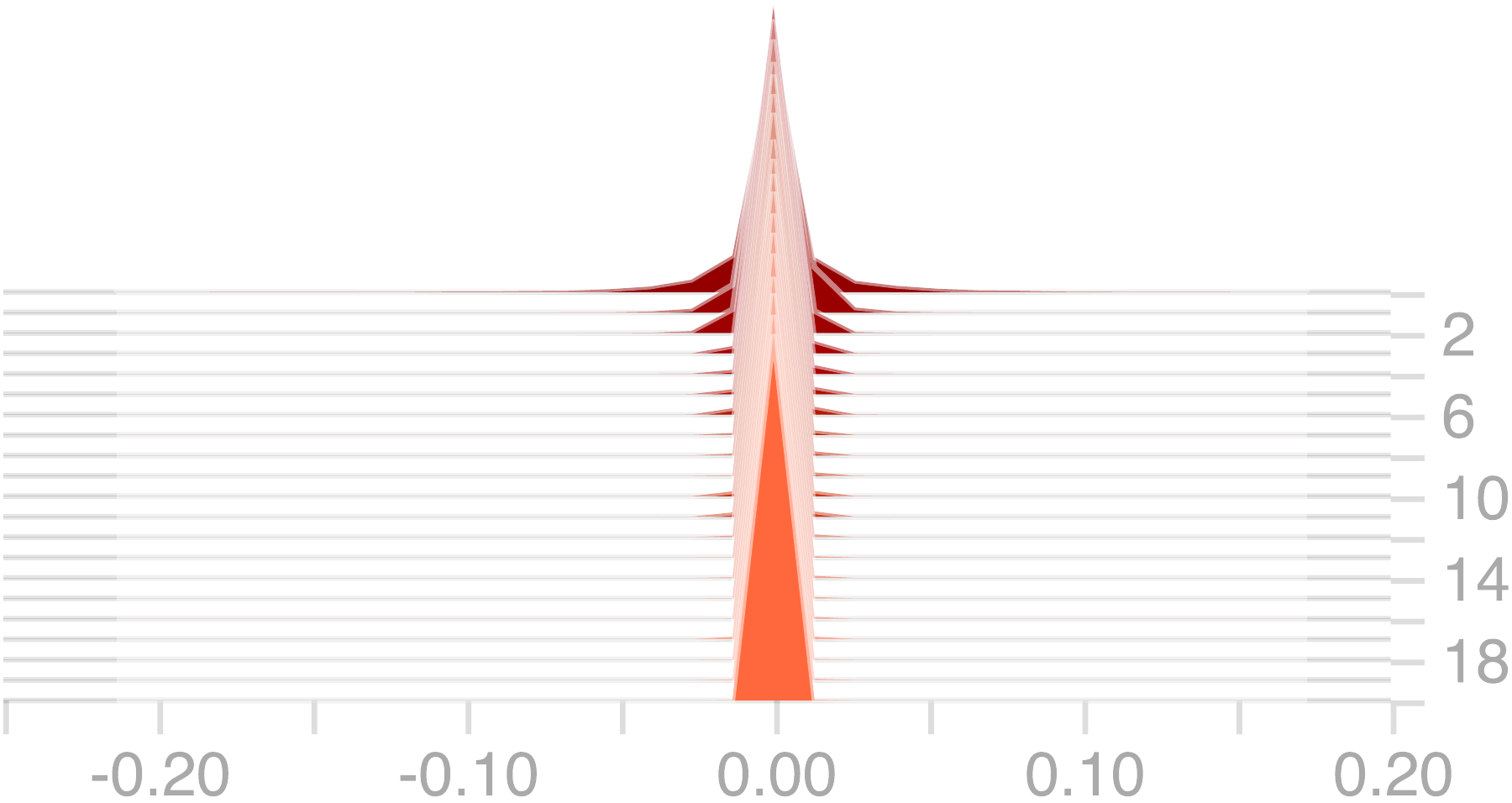}}
  \subcaptionbox{{\fontfamily{phv}\selectfont
    \footnotesize{Layer3: Dense}}}[.24\linewidth][c]{%
    \includegraphics[width=\linewidth]{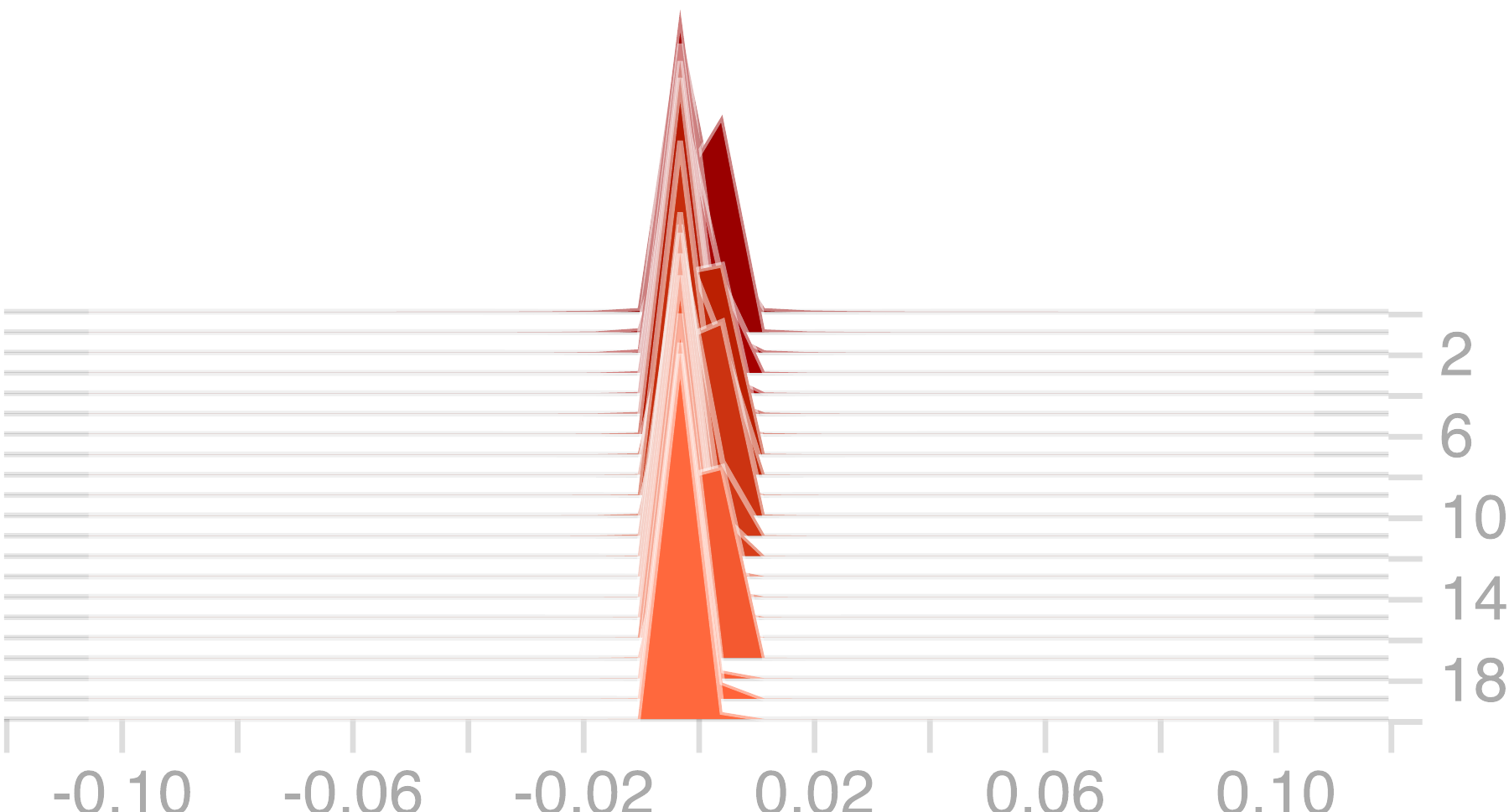}}
    \subcaptionbox{{\fontfamily{phv}\selectfont
    \footnotesize{Layer4: Dense}}}[.24\linewidth][c]{%
    \includegraphics[width=\linewidth]{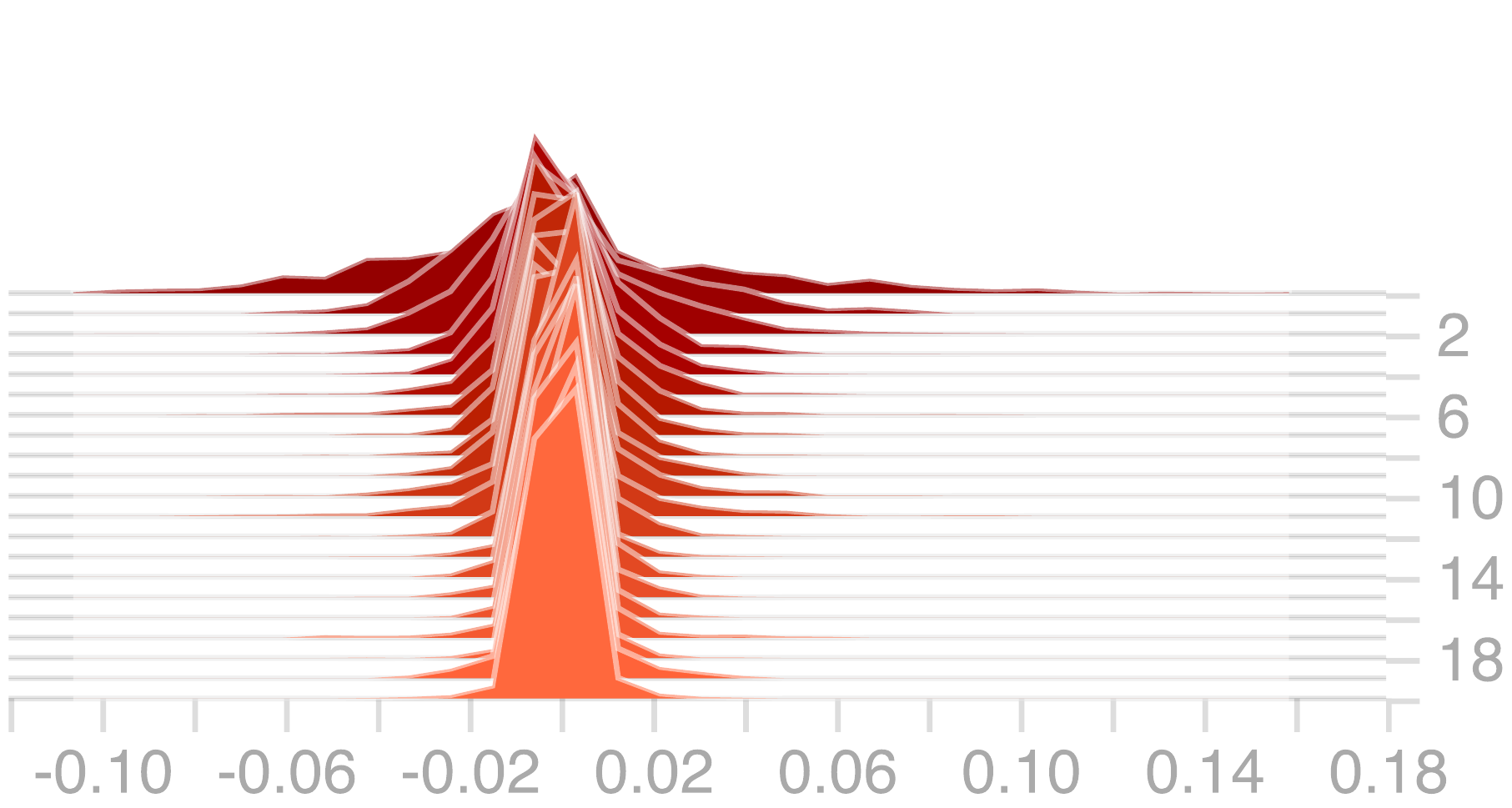}}
  \caption{Weights of a four-layer CNN (we will describe this network later in the Experiments section). The top row presents FedZip weights in different layers, during 20 rounds of training FL for a client, and the bottom row presents similar weights in different layers for the FedAvg framework. The differences in terms of  sparsification and quantization into three clusters is self-explanatory in our framework.}
  \label{fig:fedzipfedavg}
\end{figure*}



Table \ref{table:overview} summarizes a comparison between our framework and state-of-the-art compression methods from four different perspectives: compression rate (CR), robustness when handling non-iid datasets, auxiliary techniques, and the direction of the compression. The use of auxiliary techniques adds to the complexity of implementation; frameworks without auxiliary techniques are easier to adapt and faster to run. Besides, auxiliary techniques even impose additional time for preprocessing the model, such as warm-up time, fine-tuning, etc.

The term ``Upstream'' in Table \ref{table:overview} refers to sending compressed updates (including $\Delta w$) from clients to a server, while the term ``Bidirectional'' refers to sending compressed updates between a server and clients in both directions. 
\section{Problem Statement}

\subsection{Federated Parameters}
As has been previously proposed~\cite{McMahan17a}, three variables are involved in the construction of FL frameworks: 
\begin{enumerate}
\item{The random selection of clients, which is controlled by parameter $C$. This is determined on the server with respect to the bandwidth capacity and the number of clients involved.}
\item{The number of training passes each client makes over its local dataset on each round, is defined by $E$ (stands for local Epoch).} 
\item{The local mini-batch size, which is defined by $B$.} 
\end{enumerate}

A value of $C=1$ means all clients contribute to the process of averaging in the server, $C<1$ means a subset of clients is randomly selected to contribute, and $C=0.0$ means one client contributes at a time. The number of training passes (Epoch) each client makes over its local dataset is defined by $E$. Element $B$ is the local mini-batch size and describes the amount of the local dataset that is covered. $B=\infty$ means all data points are fed to the model. Assuming $x_i$ is the data point, and $y_i$ the label for a particular data point, $i$ is their index in the datapoint, and $w$ represents the weights and biases of the neural network model. 

In both the FedAvg and FedSGD frameworks, a common supervised objective function is suggested by McMahan et al.~\cite{McMahan17a}: $f_i(w) = l(x_i, y_i, w)$, a loss function for prediction on the data point $(x_i, y_i)$.
The first line of Equation \ref{eq-1} is a \emph{distributed optimization model}, and its objective is to minimize the loss function $F_m(w)$. The function $F_m(w)$ is the local objective optimization in each client. Assuming there are $M$ clients, if the $m$th client has $p$ data points, $ |p| = n_m $ is the number of data points that belong to client $m$; they are a fraction of the whole dataset with size $n$.
\begin{equation}
  \begin{split}
      \underset{w \in \mathbb{R}^d}{\min}\: \textit{f(}w) &\quad \textrm{where}\quad  {f(w) = \sum_{m=0}^{M}{\frac{n_m}{n}}\times{F_m (w)}} \\
      &\textrm{with} \quad{F_m (w)= \frac{1}{n_m} \sum_{i}^{n_m} f_i(w)}\\
  \end{split} 
 \label{eq-1}
\end{equation}
Equation \ref{eq-2} formalizes the \emph{global averaging process}. If we consider ($B$=$\infty$ , $E=1$), we have FedSGD, which averages the gradients in each round on the server. As a more generalized approach, FedAvg performs one or more iterations ($E$) on the local dataset and averages delta weights during local training. In FedAvg, we consider the element $B$ as a finite batch (e.g. $B=32$ data points) to have more than one or more iterations. The global learning rate, $\eta$, determines the effect of the weights at time $t$ on the global model's weight. We used a global learning rate $\eta=0.25$ for FedAvg and FedZip, and $\eta=0.6$ for FedSGD which is experimented in \cite{McMahan17a, McMahan17b}. In Equation \ref{eq-2}, $\eta $ = 1 means a weighted average based on the number of used data points in each client divided by the total number of data points for all clients.

\begin{equation}
\begin{split}
  w_{t+1} \leftarrow & {w_{t} - \eta \sum_{m \in S_t} \frac{n_m}{n}  (w_{t}-w_{t+1,m})}\\
\end{split} 
\label{eq-2}
\end{equation}

The data points ($i$) are partitioned over $m$ clients as a non-iid setting; $S_t$, a subset of $m$ clients, is randomly selected in round $t$, and subscript $m$ refers to the clients in this subset at round $t$ (the number of clients in the $|S_t| = m $, is controlled by element $C$).
Equation \ref{eq-1} formalizes the general FL objective function. We refer to $(w_{t}- w_{t+1,m})$ as $\Delta w$ for each client throughout this paper; $\Delta w = $\{$\Delta w^1, \Delta w^2, \Delta w^3 ..., \Delta w^m$\} are the weights and biases of the neural network models used for training in the subset of $m$ clients which post the update message from each client to the server. Thus, $\Delta w_t$ is the global weight averaged from all clients at round $t$. 

\subsection{Client Data Constraints}
Based on the nature of ubiquitous device datasets, and their distribution over clients, it is important to determine the  data attributes for efficient transformation. Therefore, we assume a local dataset resides on the client device and the connection has the following attributes \cite{McMahan17a}:
\begin{itemize}
 \item {\emph{Non-IID}}: the distribution of the data over clients is not independent and identically distributed.
  \item {\emph{Unbalanced}}: The number of data points in each client is not necessarily equal.
  \item {\emph{Limited communication}}: The connection between clients and servers is expensive, slow, and limited.
  \item {\emph{Non-convex problem}}: there are multiple local optima, which makes identifying  the global optima complex.
\end{itemize}

\subsection{Compression Parameter}
Sattler et al. \cite{Sattler18} formalize the total bit usage ($b_\textit{total} $) in the communication procedure for training a model. Let $N$ be the number of training rounds, $f$ be the frequency of communication (inverse relation with $E$), and $b$ ($|\Delta w_{\neq0} |$ ) be the number of bits that are transferred. $C$ is the fraction of clients who performed training on their dataset and their delta weights aggregated on the server. Then, equation \ref{eq:compression} presents the \emph{computational complexity} of the compression parameter based on bit utilization ($b_{total}$).
\begin{equation}
 b_{total} \in O(N \times f \times b \times C) \label{eq:compression}
\end{equation} 

As Equation \ref{eq:compression} suggests and also mentioned by Sattler et al.~\cite{Sattler19}, there are three general approaches for compression (reducing $b_{total}$): 
\begin{itemize}
  \item Reduce communication frequency $f$.
  \item Reduce the entropy of the weight updates ($\Delta w$) by decreasing the $b$ element in Equation \ref{eq:compression}.
  \item Use more efficient encoding to communicate the weight updates, thus reducing the global learning rate $\eta$.
\end{itemize}
FedZip focuses on the second and third approaches to reduce the entropy of $\Delta w $ and encodes the communication update. Note that if we reduce the communication frequency by some specific factor, the compression rate will increase by that particular factor based on Equation \ref{eq:compression}; this is not efficient from a communication perspective.

We can summarize the problem that our framework handles as follows:

\textbf{Problem:} \emph{Assuming $N$, $E$, $b$ and $C$ are known and configurable (in Equation \ref{eq:compression}), given 
that we do not increase $E$ or $N$, the objective of FedZip is to reduce $b_{total}$, while maintaining the convergence speed and accuracy of baseline FL.}

\section{FedZip Framework}
A major novelty of our framework is compressing bits per communication round (see  Equation~\ref{eq:compression}).
Other elements in Equation \ref{eq:compression}, such as $f$, can be manipulated easier than $b$ because they are not related to models' infrastructure. 
Each layer in the model stores its values separately. The ``exchanged data'' updates clients and server illustrate with $\Delta w$. Each value in $\Delta w$ is stored as a default in a 32-bit float. In the first step, FedZip sparsifies $\Delta w$, and in the second step, it quantizes $\Delta w$.
Sparsification and quantization modules will be described in more detail.

Note that each layer of the model is analyzed as a separate tensor by our modules in this framework. For instance, a simple four-layer neural network includes four bias tensors and four weights tensors. FedZip feeds each tensor to the related module separately (Figure \ref{fig:fedzipfedavg}).

\subsection{Sparsification}
Sparsification is used to prune the weights to increase communication efficiency. Tsuzuku et al.~\cite{Tsuzuku18} use sparsification to pass the highest gradient \footnote{In this context, the gradient refers changes between current weights and biases after back-propagation that the error is applied on weights and biases.} to the server when the element magnitude of gradients is greater than an arbitrary threshold.

To implement sparsification in distributed learning, a fixed rate, such as $90$ percent (employed by Aji et al.~\cite{Aji17}) is used. QSGD \cite{Alistarh17} focuses on using a lower bound for sparsity, i.e. if a magnitude of a value in $\Delta w$ is lower than a specific rate, then the value will be removed from the tensor. The authors of QSGD \cite{Alistarh17} claim that sparsification does not significantly affect convergence and the effect will be trivial. However, we use a Top\_$z$ sparsification method \cite{Aji17} among others, because it will convert lower magnitudes (pruning in Top\_$z$ does not affect the inference in a neural network model) to zero and reduce the variance of $\Delta w$. In particular, first, a subset of values with top magnitude will be identified. Then, each value in each tensor will be presented with $w_i$ (where $i$ is the $i$th value in a flattened tensor) and $w_{z}$ is the $z$th greatest value in the corresponding tensor $\Delta w$. 
Equation \ref{eq:spars} formalizes our sparsification approach.

\begin{equation}
Top\_z(w)=
\begin{cases}
w_i,\quad\enspace if \quad w_i>w_z \\
0,\quad\quad if \quad  w_i<w_z
\end{cases}
\label{eq:spars}
\end{equation}

Our sparsification module is applied separately to each layer, which allows our framework to treat sparsity as a user-selected parameter. It can change for each tensor based on the number of neural network parameters. After several experiments, we realized that sparsifying the bias tensor with a lower rate (e.g. less than Top-$0.2$) allows it to reach convergence more quickly. Reporting the details of this experiment is not in the scope of this paper, but, it is notable that Bhattacharya and Lane~\cite{Bhattach} suggest setting the bias of each node to zero and abandoning this parameter. 

\subsection{Quantization}

After sparsifying $\Delta w$, FedZip quantizes each tensor in $\Delta w$. We use a statistical quantization form, in which each tensor includes biases and weights (separately for each layer of the model) is quantized by a $k$-mean clustering algorithm. We use clustering because the centroids of each cluster offer a reasonable reflection of the distribution for the tensor’s value, rather than other quantization methods, in which a fixed or stochastic set is selected. In other words, there are methods (e.g. STC \cite{Sattler18}, SignSgd \cite{Bernstein18}, and TernGrad \cite{Wen17}) that do not consider the distribution of weights. Our experiments and comparison with other methods in \ref{tab:cmpDiffApp} show that not taking into account the distribution of weights causes skewing a learning trajectory in the optimization space, leads to high variance, and an increase in training rounds. Therefore, those quantization methods do not introduce the best representative values of a tensor.
In contrast, the use of $k$-means clustering for quantization avoids this by converting the value of a tensor to a number of clusters and using the centroid of those clusters as the best representative. Later in our experiments, we empirically demonstrate that this special sequence of sparsification, quantization, and encoding reach baseline accuracy with the same convergence speed.

To identify the optimal number of clusters, we use the silhouette index and identify $k=3$ as the best value for the $k$-mean hyper-parameter for both datasets. The centroid in Equation \ref{eq:kmeans} is represented by $c$ where $c_j$ is the $j$th predicted centroid.
The number of clusters ($k$) directly affects the compression rate. Thus, there is a trade-off between the number of clusters and the compression rate that affects accuracy. The quantization reduces the variance of weights by substituting all the values to three centroids values. As the result, the encoding will be smaller due to the decreased number of unique codes required to encode $\Delta w$ (Equation \ref{eq:kmeans}).

\begin{equation}
\begin{gathered}
    \textit{k-means}(\Delta{w})  =  \{\forall w_i \in \Delta{w}_m, \quad let \quad w_i \leftarrow c_j\}
\end{gathered}
\label{eq:kmeans}
\end{equation}

\begin{algorithm}[htbp]
\caption{FedZip. The clients are indexed by $m$; $P_m$ is a set of data for the $m$th client, $E$ is the number of local epochs, $C_M$ is the number of all clients for FedAvg, $\eta$ is the learning rate, and $B$ is the local mini-batch size.}\label{euclid}

\begin{algorithmic}[1] 
\Procedure{Server Execution:}{}
\label{alg:FedZip}
\State $\text{initialize}\enspace w_{t=0}$
\For{$\text{round}\enspace \textit{t = 1,2,... }$} 
  \State $m \leftarrow max(C.M, 1)$
  \State $S_t \leftarrow \text{(random set of $m$ clients)}$

  \For{$\text{client}\enspace m\text{th}\in S_t \;  \textit{in parallel}$} 
  \State $msg^m_{t+1}, \theta^m_{t+1} \leftarrow ClientUpdate(m, w_t)$
  \State $w^m_{t+1} \leftarrow decoding(msg^m_{t+1}, \theta^m_{t+1})$ 
\EndFor
\EndFor
\State $w_{t+1} \leftarrow \sum_{m=1}^{M} \frac{n_m}{n} w^m_{t+1}$
\EndProcedure 

\Procedure{\text{Client Update(m, w):}}{}
\State $B \leftarrow (\text{split}\enspace P_m\enspace \text{into batches of size}\enspace B) $
\For{$\text{local epoch}\enspace i\enspace \text{from 1 to E}$} 
  \For{$\text{batch } b \in B $} 
      \State $ w \leftarrow w \: - \eta \, \nabla \, l(w, b)$
  \EndFor  
\EndFor
\State $ \textbf{encoding} (\Delta w): $
     \State \quad\quad $msg^m_{t+1}, \theta^m_{t+1}  \leftarrow encoding(w)$
\State \Return {$\enspace msg^m_{t+1}, \theta^m_{t+1} \enspace \text{to the Server}$}
\EndProcedure
\end{algorithmic}
\end{algorithm}


\begin{algorithm}[htbp]
\caption{Encoding framework; $c_j $ refers to the $j$th centroid among clusters. As it is mentioned, $|c_j|=3$ means we have three centroids.}
\label{alg:compression}
\begin{algorithmic}[1]
\Procedure{Encoding:}{}
\State $\textbf{Sparsification}$
\State \hskip1.5em $ w \leftarrow top{\text -}z(w)$
\State $\textbf{Quantization}$
\State \hskip1.5em $ w \leftarrow K{\text -}means(w) $
\For{$ \text{each Centroid of } w_m \in w$}
    \State $ w_i \leftarrow c_j $
\EndFor
\State $\textbf{Encoding}$
\State \textit{select one of the methods below to build\newline 
\quad\quad the update message}
\State \hskip1.5em $ \text{1-} msg^m_{t+1}, \theta \leftarrow \text{Huffman}(w)$
\State \hskip1.5em $\text{2-} msg^m_{t+1}, \theta \leftarrow \text{Exact Position}(w)$
\State \hskip1.5em $\text{3-} msg^m_{t+1}, \theta \leftarrow \text{Difference of Address Position}(w)$\footnotemark
\State \textit{send the decoding table}
\EndProcedure
\Procedure{Server:}{}
\State $\textbf{Decoding}$
\State \hskip2.5em $ w \leftarrow \text{decoding}(msg, \:\theta)$
\EndProcedure
\end{algorithmic}
\end{algorithm}
\footnotetext{This method results in the best CR.} 

\begin{table*}[htbp]
\begin{center}
    \captionsetup{font=small,skip=10pt}
    \caption{The results of applying proposed framework with two models (CNN and VGG16) on the MNIST dataset.}
    \begin{threeparttable}
    {\small
    \begin{tabular}{|c|c|c|c|c|c|c|c|c|c|}
    \hline
    \textbf{Methods} & \textbf{Model} & \textbf{Convergence} & \textbf{Training} & \textbf{Test}  & \textbf{Loss} & \textbf{Size of Updates}  & \textbf{Compression} & \textbf{Number of} & 
    \textbf{Round (N)} \\
   
    &   & \textbf{Speed} & \textbf{Accuracy} & \textbf{Accuracy} & & \textbf{(MB)\tnote{1}}  & \textbf{Rate} & \textbf{Clients and C} & \\
    
    \hline
    FedAvg & CNN & baseline & 99.44 & \textbf{98.03} & \textbf{0.013} & 4.79 & 1x & 50, & 20 \\
    &  &  & &  &  &  & E=1, B=32 & C=1 & \\
    
    \hline
    FedSGD & CNN & low  & 99.04 & 97.65 & 0.031 & 4.79 & 1x & 50, & 100 \\
    &  & &  &  &  &  & E=1, B=32 & C=0 & \\
    
    \hline
    FedZip &  CNN & same   & 99.34 & 97.79 & 0.026 & \textbf{0.0078} & Up to \textbf{1085}x & 50, & 20 \\
    &  & &  &  &  &  & E=1, B=32 & C=1 & \\

    \hline
    FedAvg & VGG16 & baseline  & 99.82 & \textbf{94.82} & \textbf{0.1708} &
    134.54 &  1x & 50, & 20\\
    &  &  &  &  &  &  & E=1, B=32 & C=1 & \\
 
    \hline
    FedSGD & VGG16 & low & 99.02 & 92.59 & 0.6213 &  134.54 & 1x & 50, & 100 \\
    &   &  &  &  &  &  & E=1, B=32 & C=0 & \\
    \hline
    FedZip & VGG16 & same & 99.42 & 93.30 & 0.5719 &  \textbf{0.6911} & Up to \textbf{194}x & 50, & 20 \\
    &   &  & &  &  &  & E=1, B=32 & C=1 & \\
 \hline
  \end{tabular}}
  \end{threeparttable}
   \begin{tablenotes}
  \item[1]1. Each client's size of exchanged updates is shown in this column. 
  \end{tablenotes}
  \label{tab:fedzip,fedavg}
  \end{center}
\end{table*}

\subsection{Encoding}
Both sparsification and quantization process change values in $\Delta w$. At this stage, most (though not all) $\Delta w$ values are grouped into a single cluster. Figure \ref{fig:fedzipfedavg} shows the difference between weights in FedAvg and our proposed framework. Our Quantization, which is achieved through clustering, results in $\Delta w$ represented by three centroids; each element is clustered and represented by one of the centroids.

Based on these settings, we employ ``Huffman Encoding'' \cite{Huffman52} to create an address table. Huffman Encoding assigns the shortest number of bits to frequent values, and a longer number of bits to infrequent values. In our case also, the most frequent cluster receives the shortest number of bits. 

As described in the quantization section, silhouette indexing recommends $k=3$ as the best possible number of clusters. Huffman Encoding represents the most frequent clusters with one bit and the other clusters with two bits. With Huffman Encoding, our framework reaches a CR close to $32\times$. Because of the huge number of parameters that a neural network imposes, this CR is still not sufficient. Hence, inspired by Huffman Encoding, we employ two kinds of address tables to index two infrequent clusters by their positions. The first method specifies the position of the two least frequent clusters in the $\Delta w$ and allocates them a single bit in the address table to distinguish those two clusters from each other. The second address table includes two infrequent clusters, but instead of their positions, it stores the difference between positions in the flattened $\Delta w$. 

Equation \ref{eq:encoding} presents our encoding approaches with a parameter which specificies the mode of encoding. Each of these encoding modes is optimized in a separate experiment to achieve the best CR. FedZip does not use any auxiliary techniques such as warm-up training \cite{Goyal17} or momentum correction \cite{Qian} to reach the best possible accuracy and convergence speed. It means that the FedZip has the advantage of not having the overhead of auxiliary techniques than approaches that require auxiliary techniques.

\begin{equation}
    \{ \forall w_i\in \Delta w_m  \quad do \quad Encoding{(\Delta w, \theta)}\} 
    \label{eq:encoding}
\end{equation}
The $\theta$ is a hyper-parameter of the encoding function. It indicates the method used for encoding and decoding table. There are three types of encoding implemented by our approach (i)storing Huffman Encoding in address table, (ii) Storing Address Position in address table, and (iii) Storing Differences of Address Position in address table. One of these have to be selected to encode. The best compression rate belongs to the third one, Differences of Address Position.

\subsection{Algorithms}
Based on Equation \ref{eq-1} and our proposed modules, Algorithm \ref{alg:FedZip},  describes our approach in pseudocode.
On the server, in the first round, the model is initialized with random weights (line 2). Then, the server selects a random set of clients ($S_t$) based on the fraction of clients $C$ (line 5) to broadcast the initialized weights of the defined model to the clients. From line 7, each of the selected clients starts to run their model based on the local dataset and creates an update ($\Delta w$).

\begin{figure*}[htbp]
\centering
\includegraphics[width=\linewidth]{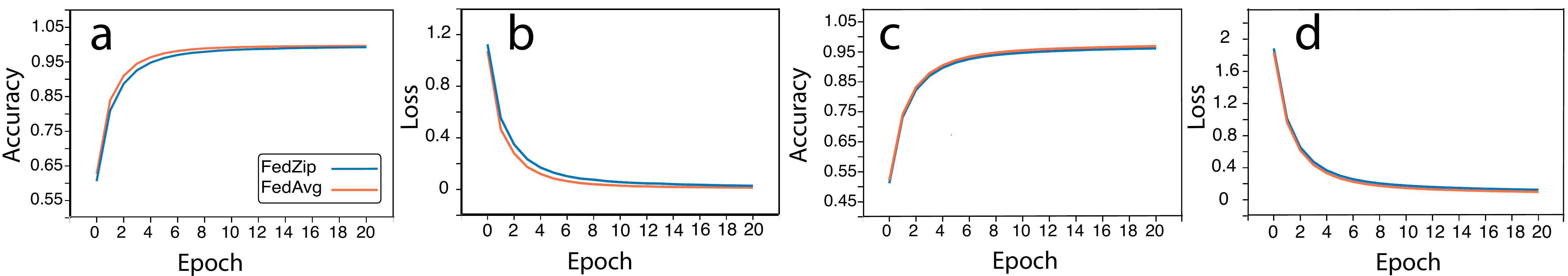}
\caption{Convergence speed of FedZip (blue lines), state-of-the-art FedAvg (orange lines) on the MNIST and EMNIST datasets, using CNN. The accuracy is illustrated in (a) and (c) and loss in (b) and (d). (a) and (b) show the MNIST dataset with 10 classes, while (c) and (d) show the version with 62 classes (EMNIST).}
\label{fig:convergence_me}
\end{figure*}

Next, in line 8, the encoded module decodes the compressed message for time $t$ for client $m$ and prepares the $\Delta w$ for each client to be aggregated on the server. In line 11, the aggregated $\Delta w$ is averaged on the server to build the global weights and biases as a start point for the next round. The ``clientUpdate'' procedure occurs for each client in parallel (line 7). On the clients, a batch size proportional to the local dataset ($P_m$) must first be determined (line 14).
Next, $\Delta w$ is calculated for each client after training $E$ times on the local dataset (line 17), with a coefficient $\eta$. Then, FedZip applies lossless compression (line 20, 21) on all clients’ delta weights in order to build the update message to return to the server (line 17) from the client. The training is performed in parallel due to the synchronous architecture of FL. The compression framework, which is mentioned as the encoding and decoding in Algorithm \ref{alg:FedZip} (line 8, 21), is described in detail in Algorithm \ref{alg:compression}.

In Algorithm \ref{alg:compression}, we provide the sequence of encoding procedure happening on the clients. There are three methods to build our compression framework. Sparsification (line 3), Quantization (line 5), and compression (line 10) and sending the compressed $\Delta w$ and address table ($\theta$) (line 14). On the server, the decoding procedure happens on the server based on address table ($\theta$), Centroids $C_j$, and $\Delta w$ (line 18). 
\section{Experiments}\label{experiments}
In this section, we first describe two datasets that we have used for our experimental evaluation, including a baseline dataset used for deep learning experiments and a dataset with data from wearable devices. Then, we compare the accuracy of our approach to state-of-the-art approaches. Next, we discuss the compression impact of FedZip and its scalability. Finally, we report the impact of implementing FedZip on smartphones and its energy utilization and response time with state-of-the-art FL frameworks. \footnote{To allow for the full reproducibility of our results, all of our code is available in a GitHub repository. [https://github.com/malekijoo/FedZip]}

\subsection{Dataset}
We evaluate our framework on two real-world datasets. The first is MNIST \cite{LeCun18}, a benchmark dataset for deep learning algorithms \cite{TensorFlow}. It is based on LEAF \cite{Caldas18}, but partitioned especially for FL, with two different forms of image: only-digit MNIST (10 classes) and with letter EMNIST (62 classes).

The second dataset is the ``insight4wear'' dataset \cite{reza2015}, made up of data from a multi-sensor smartwatch.\footnote{ This dataset includes information on heart rate, user interaction with the screen of the watch, battery status (i.e. charging or discharging), notifications arriving on the watch screen, and Bluetooth connection of the watch to the phone (i.e. when the watch is connected and disconnected from the phone).} We decided to use this dataset to challenge our proposed framework with real-world data obtained from wearable devices because we believe that FL will be integrated into wearables in the near future.

\begin{figure}[htbp]
\centering
\includegraphics[width=0.5\linewidth]{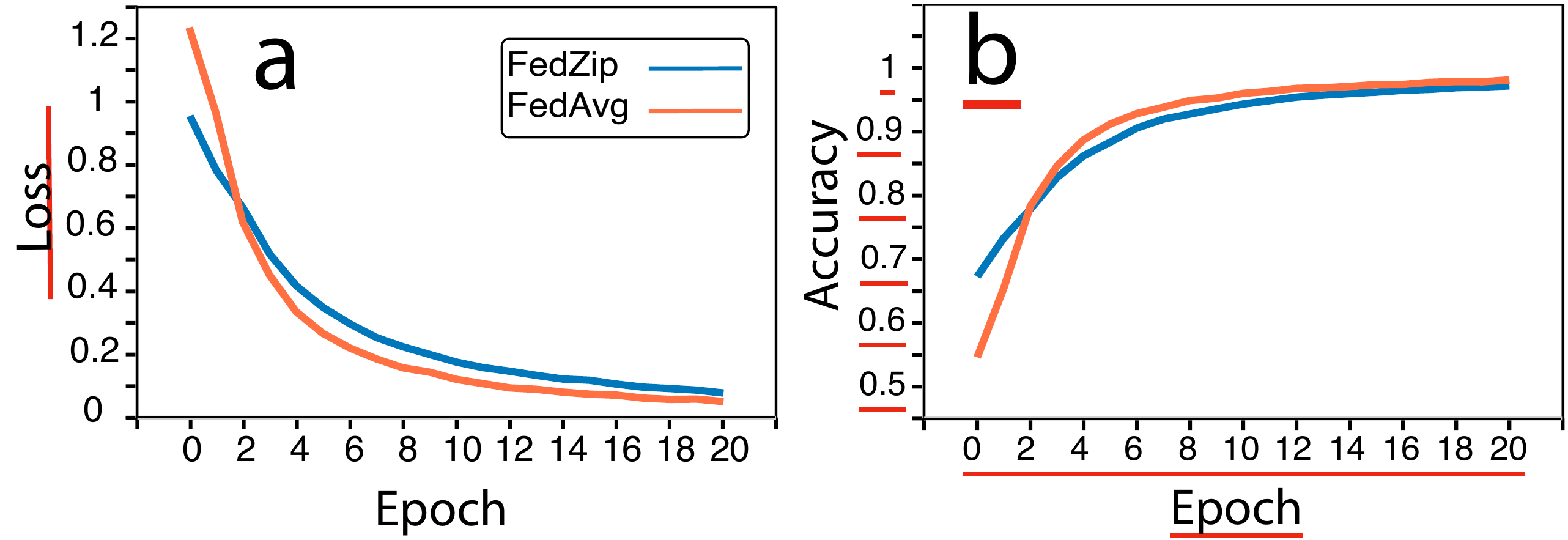}
\caption{Training convergence speed of VGG16 model using FedZip (blue lines) and state-of-the-art FedAvg (orange lines), on the MNIST dataset. The loss is illustrated in (a) and accuracy in (b) over 20 rounds.}
\label{fig:convergence_vgg}
\end{figure}
\begin{figure}[htbp]
\centering
\includegraphics[width=0.5\linewidth]{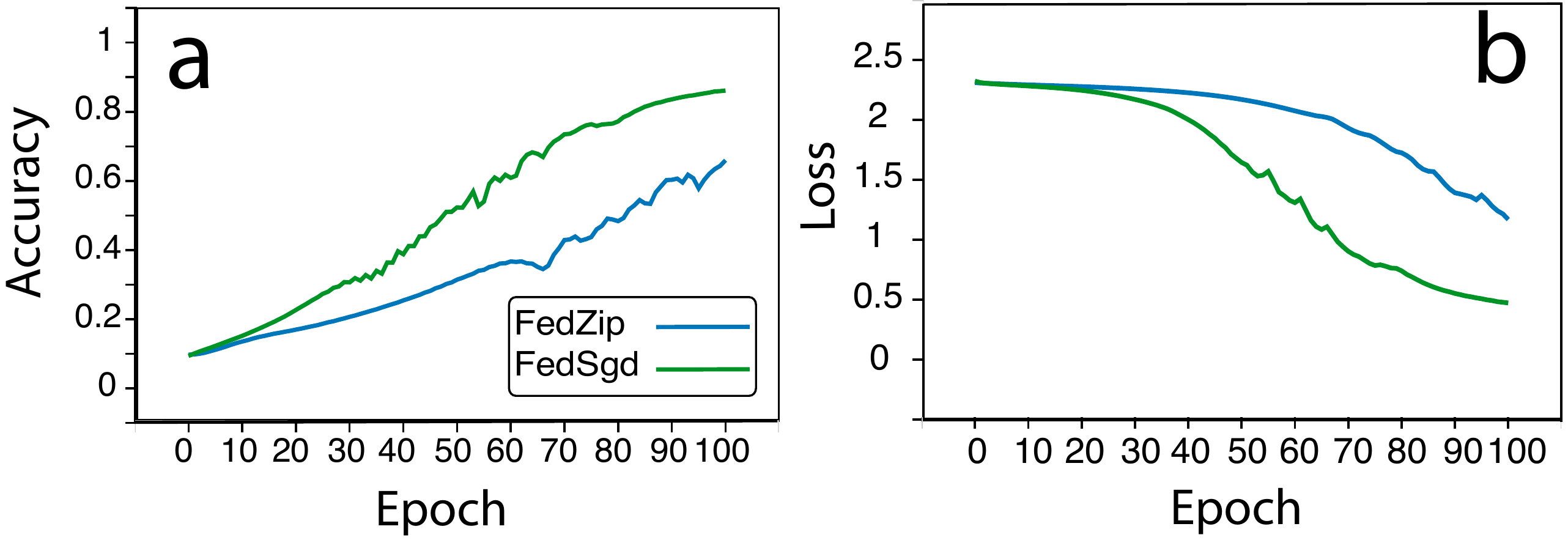}
\caption{Convergence speed of FedZip (FedSGD with Compression, blue lines), FedSGD (green lines), on MNIST dataset. The accuracy is illustrated in (a) and loss in (b) for 20 rounds. The result is showing the evident effect of compression on accuracy and loss degradation in SGD model.}
\label{fig:convergence_sgd}
\end{figure}

Homayounfar et al. \cite{morteza20} have built a fully convolutional neural network (FCNN) which is used to quantify battery drain symptoms by classifying smartwatch events into high vs. low battery use via binary classification. Their model has 4.928 million parameters, and this FCNN is specifically designed for battery-powered devices, such as mobile phones and wearables.

\subsection{Accuracy}
The most important factor  affecting the compression is the accuracy of the classification. Thus, first, we report the accuracy. 
\subsubsection{Experiment on MNIST and EMNIST datasets}
To conduct experiments on MNIST dataset (digits only) and EMNIST dataset (digits and letters), we applied a simple four-layer convolutional neural network including two layers of the convolution and two layers of the dense layer. We refer to this model as CNN in the rest of the paper. 
The FedZip framework obtains train and test accuracy similar to the FedAvg baseline. We report a comparison between the accuracy of FedZip and FedAvg (a state-of-the-art algorithm designed for FL architecture)---similar approaches applied to the same dataset---in Table \ref{tab:fedzip,fedavg}.
The convergence speed is illustrated in Figures \ref{fig:convergence_me} to \ref{fig:convergence_sgd}. Though Figures \ref{fig:convergence_me}(a, c), \ref{fig:convergence_vgg}(a), and \ref{fig:convergence_sgd}(a) show that our framework does not affect accuracy, the curves of FedZip are similar to those of FedAvg for the MNIST dataset and converge at 20 rounds.

In addition to the CNN model, we deploy the experiment employing the Very Deep Convolutional Networks for Large-Scale Image Recognition (VGG16)  model \cite{Karen14} for MNIST. The convergence is illustrated in Figure \ref{fig:convergence_vgg}(a, b). The VGG16 model consists of 13 convolution layers and ended with three dense layers. It has about $34$ million parameters. Since each parameter is stored in a 32-bit float, the model occupies 134 MB. In other words, if no FL algorithm is used, every single device that uses VGG16 must transfer 134 MB of data to the server and receive it back in each execution, which is extremely expensive from a network usage perspective. 

We conduct another experiment and compare our proposed compression framework with FedSGD (another state-of-the-art FL compression). Based on the nature of the SGD algorithm, any decrease in the fraction of clients ($C$), slows down the convergence speed and thus the accuracy degrades further. We examine the FedSGD with MNIST (10-class), which has only numbers. Figure \ref{fig:convergence_sgd}(a, b) presents the results of our comparison. The change in training accuracy between performing the compression and applying FedSGD shows the degradation. The accuracy and loss measure shows that compression implemented with FedSGD leads to less-accurate results.

Additionally, we report the CR and accuracy degradation of previous state-of-the-art methods and compare them with FedZip (Table \ref{tab:cmpDiffApp}). The accuracy degradation of our approach (FedZip) is about one percent for different shapes, parameters, and sizes of neural network models applied to different datasets.

Figures \ref{fig:convergence_me} to \ref{fig:convergence_sgd} show different experiments based on VGG16 and CNN models and FedAvg and FedSGD as state-of-the-art FL architectures. Figures \ref{fig:convergence_me}(a, c), \ref{fig:convergence_vgg}(a), and \ref{fig:convergence_sgd}(a) show the accuracy, and \ref{fig:convergence_me}(b, d), \ref{fig:convergence_vgg}(b), \ref{fig:convergence_sgd}(b) the loss during training. Figures \ref{fig:convergence_me}(a, b, c, d), and \ref{fig:convergence_sgd}(a, b) are results from the CNN model and  \ref{fig:convergence_vgg}(a, b) from the VGG16 model \cite{Karen14}. Figures \ref{fig:convergence_me}(a, b, c, d) and  \ref{fig:convergence_vgg}(a, b) show results from FedAvg, while \ref{fig:convergence_sgd}(a, b) show results incorporating the FedSGD model during learning. All the orange line in Figures \ref{fig:convergence_me} and \ref{fig:convergence_vgg} show the FedAvg baseline, the green line in Figure \ref{fig:convergence_sgd} is the FedSGD and the blue lines are our model during training rounds. FedSgd and FedZip with $C<1$ (client fraction) have slower convergence speed and reach their best accuracy in 100 rounds, however, the FedAvg and FedZip with $C=1$ have a much higher convergence speed and reach the best precision in 20 rounds.

\begin{figure*}[htbp]
  \centering
  \captionsetup[subfigure]{labelformat=empty}
  \subcaptionbox{{\fontfamily{phv}\selectfont
    \large{   a}}\label{fig:myplotcnn1}}[.32\linewidth][c]{%
    \includegraphics[width=\linewidth]{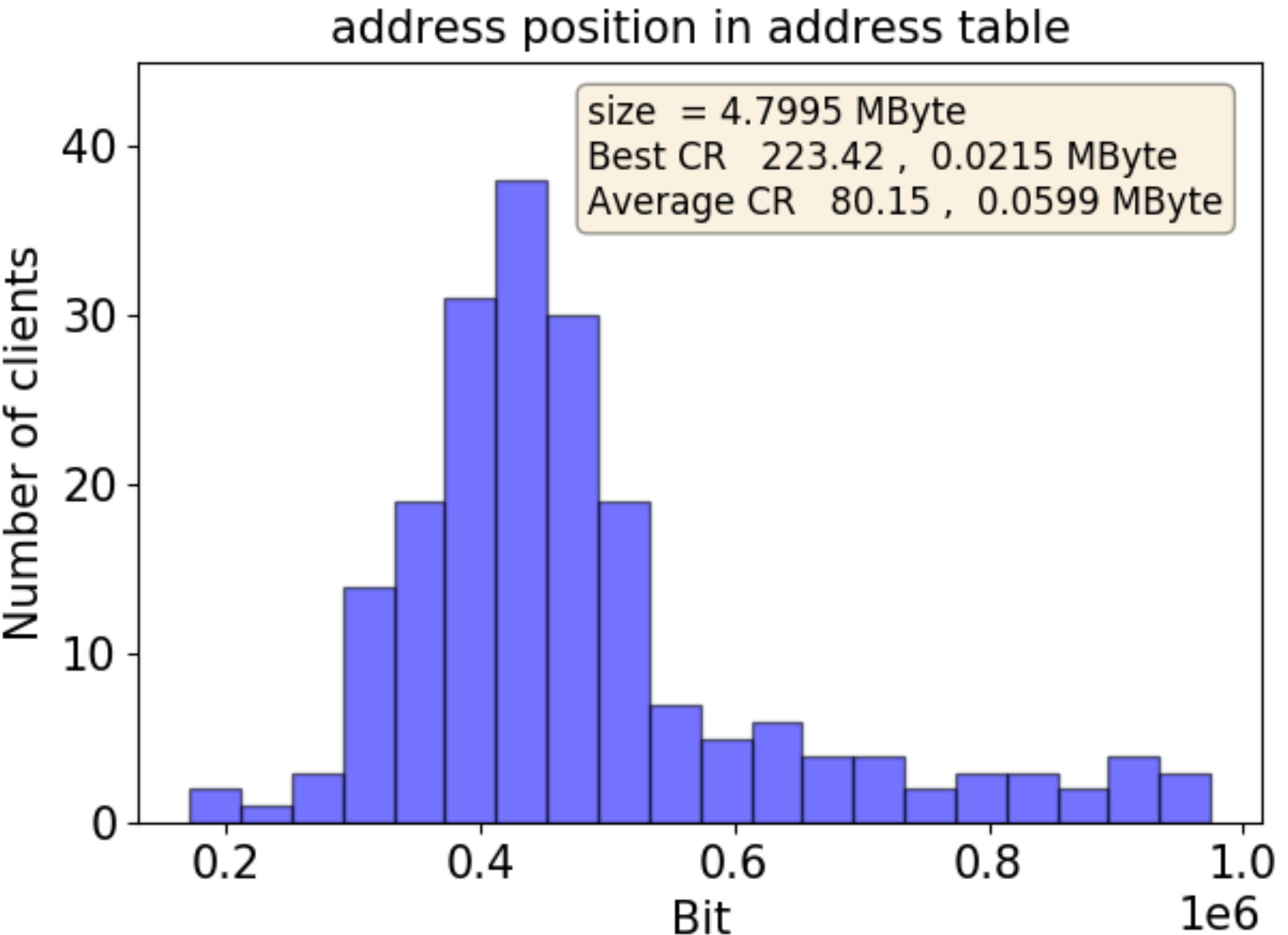}}
  \subcaptionbox{{\fontfamily{phv}\selectfont
    \large{   b}}\label{fig:myplotcnn2}}[.32\linewidth][c]{%
    \includegraphics[width=\linewidth]{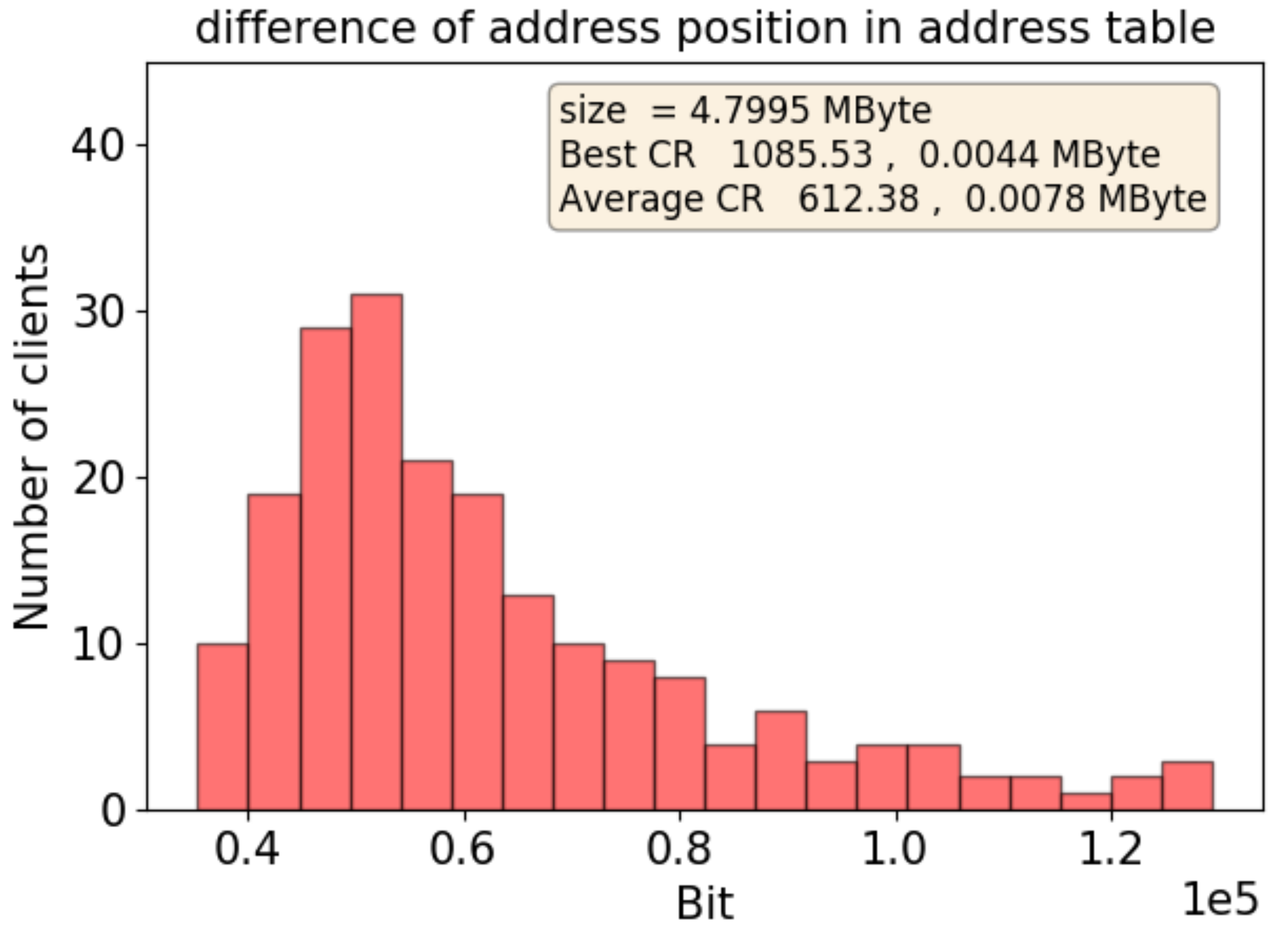}}
  \subcaptionbox{{\fontfamily{phv}\selectfont
    \large{   c}}\label{fig:myplotHuffma1}}[.32\linewidth][c]{%
    \includegraphics[width=\linewidth]{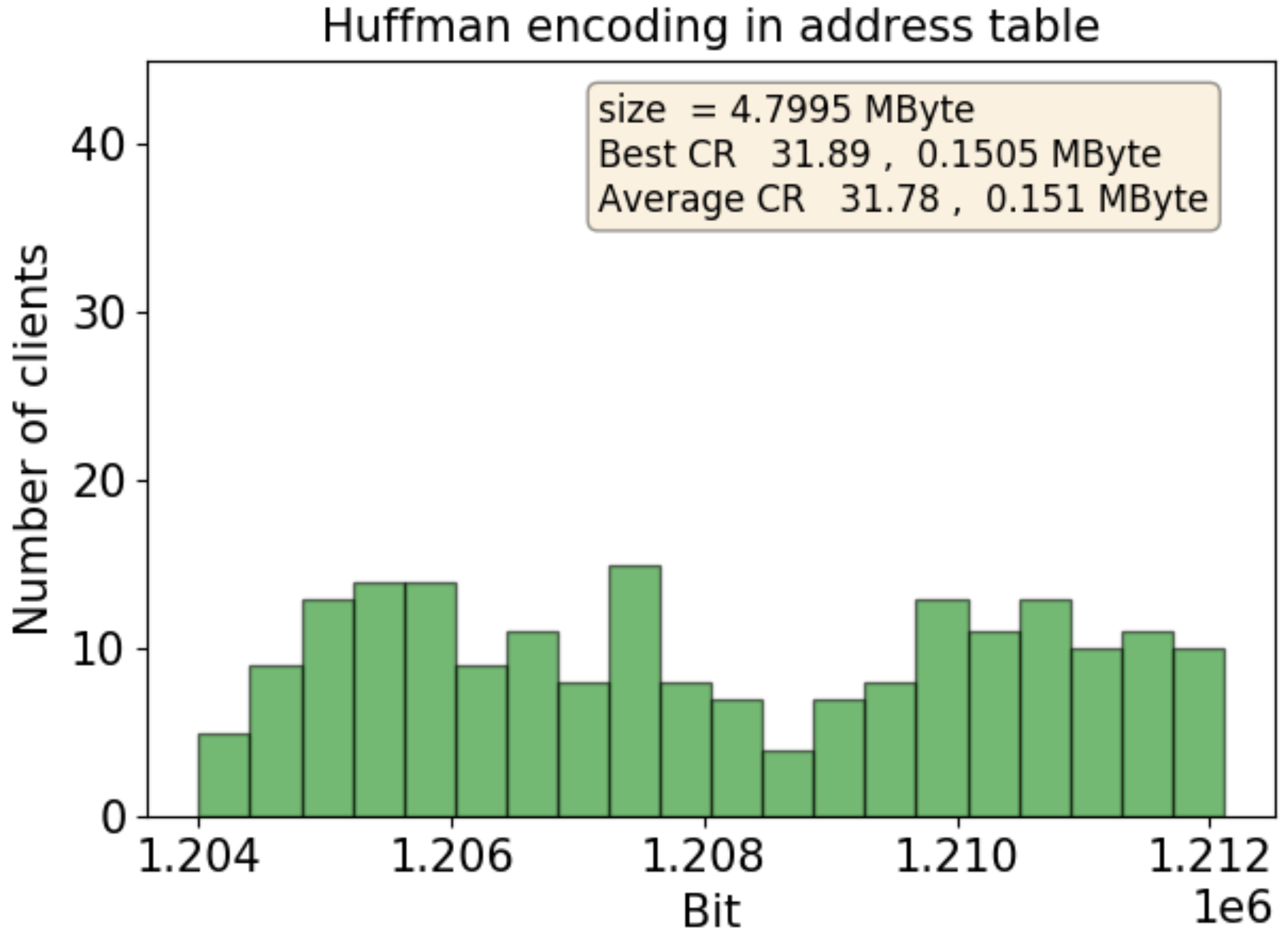}}
  \bigskip
  \subcaptionbox{{\fontfamily{phv}\selectfont
    \large{   d}}\label{fig:myplotvgg1}}[.32\linewidth][c]{%
    \includegraphics[width=\linewidth]{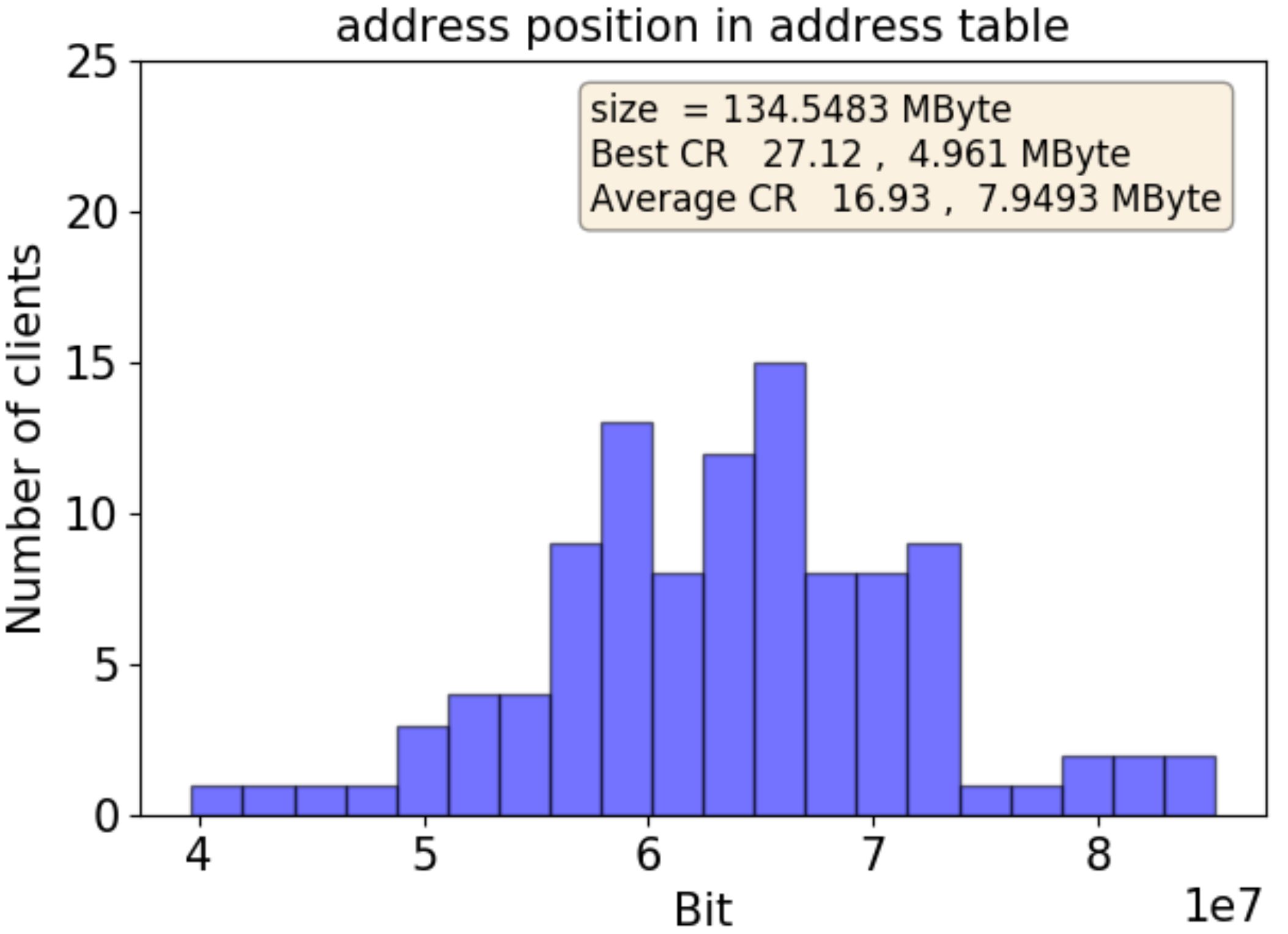}}
  \subcaptionbox{{\fontfamily{phv}\selectfont
    \large{   e}}\label{fig:myplotvgg2}}[.32\linewidth][c]{%
    \includegraphics[width=\linewidth]{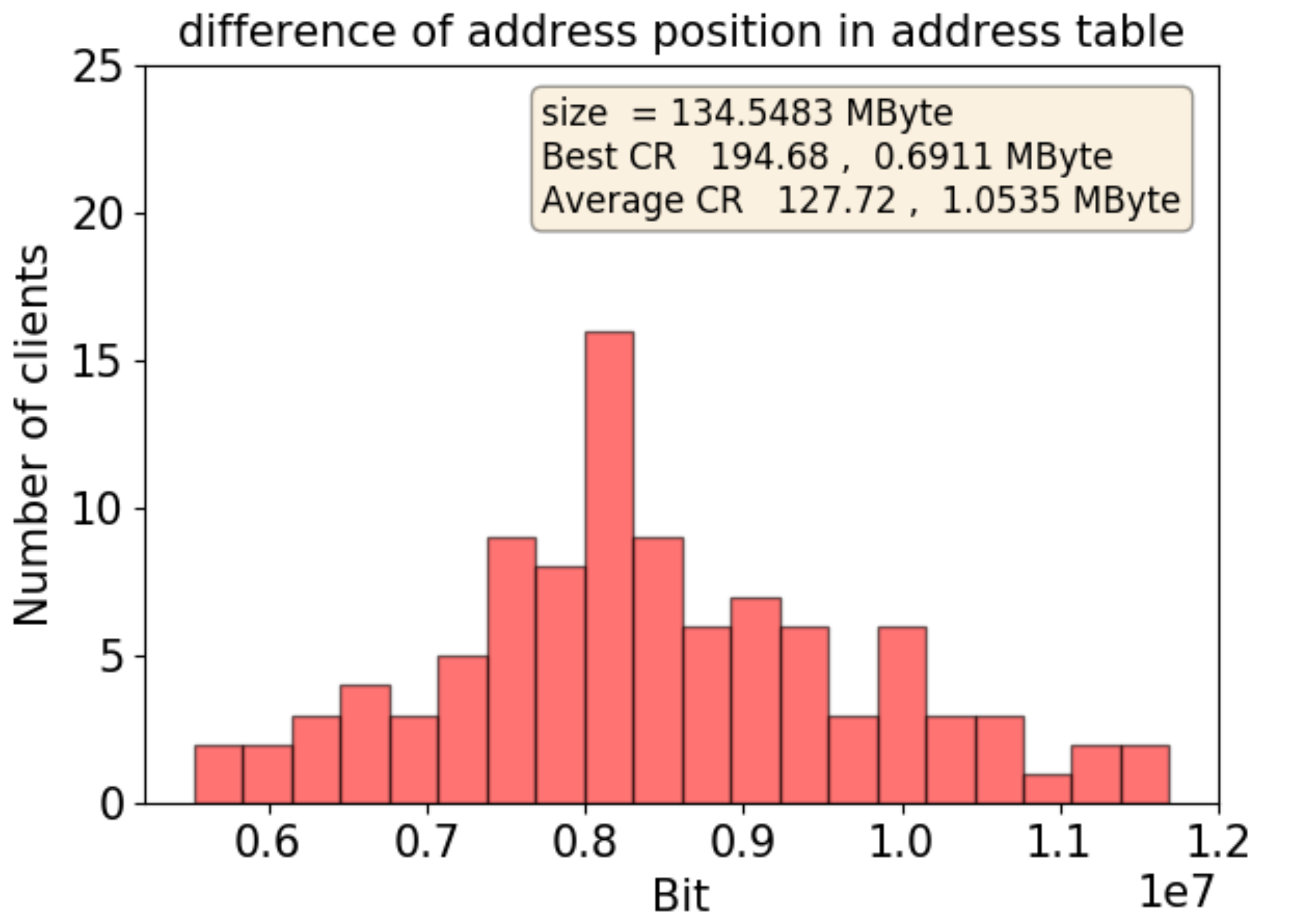}}
  \subcaptionbox{{\fontfamily{phv}\selectfont
    \large{   f}}\label{fig:myplotHuffma2}}[.32\linewidth][c]{%
    \includegraphics[width=\linewidth]{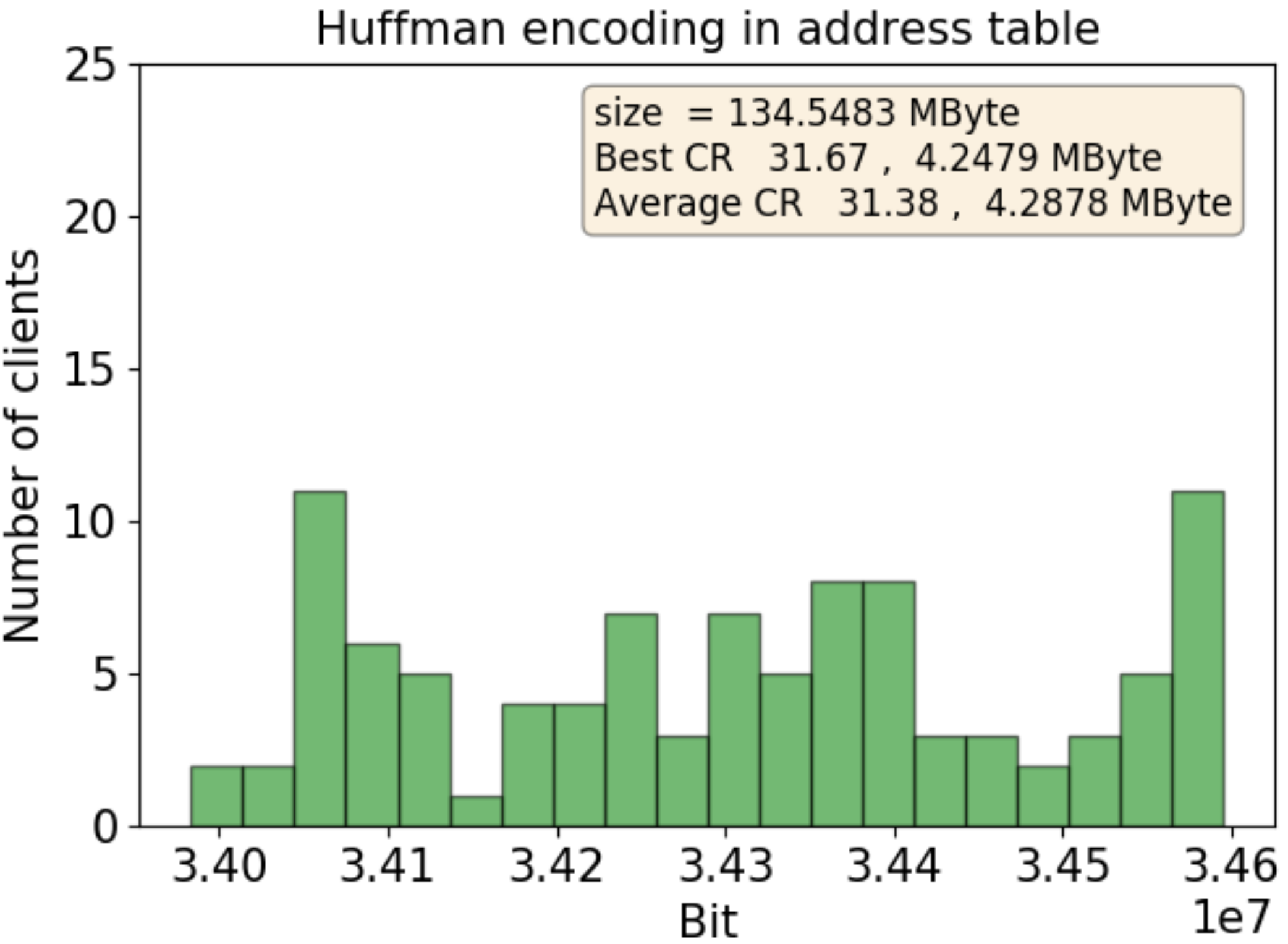}}
  \captionsetup{skip=-10pt}
  \caption{The exchanged updated sizes between server and clients, recorded during experiments, are indicated in range of bits in x-axis. Clients with approximate same update size thrown in same bin and the number of clients in each bin is shown in y-axis. (a) and (b) are histograms of 200 records of update sizes (50 clients), for CNN, with the  exact position and difference of address positions encoded in an address table.(d), (e) are histograms of 100 records of update sizes, for the VGG16 model, with the exact address position and difference of address positions encoded in an address table. (c), (f) are histograms of 200 and 100 records of update sizes, for CNN and VGG16 respectively, encoded by Huffman Encoding.}
\label{fig:histrec}
\end{figure*}

\subsubsection{Experiment on Insight4wear dataset}
We employed a model proposed by Homayounfar et al.~\cite{morteza20}. This model, which takes a centralized approach, was tested with a five-layer convolution and a cascade concatenation of those five layers, resulting in $97.82\%$ training accuracy ~\cite{morteza20}. Here, we use this model as a recently implemented example in the context of decentralized networks for wearable applications. We embed this model in each client and run FedAvg on the Insight4wear dataset \cite{reza2015}. The training phase shows a trivial decrease in loss ($ 0.089$) and accuracy ($ 97.82\%$) in 100 rounds. Next, we use FedZip (including its sparsification, quantization, and encoding methods) instead of FedAvg, resulting in $0.1213$ loss and $96.53\%$ accuracy. The test accuracy in FedZip decreases by 1\% compared to FedAvg. The results of this comparison are reported in Table \ref{tab:insight4wear}.

\begin{figure*}[htbp]
\centering
\includegraphics[width=\linewidth]{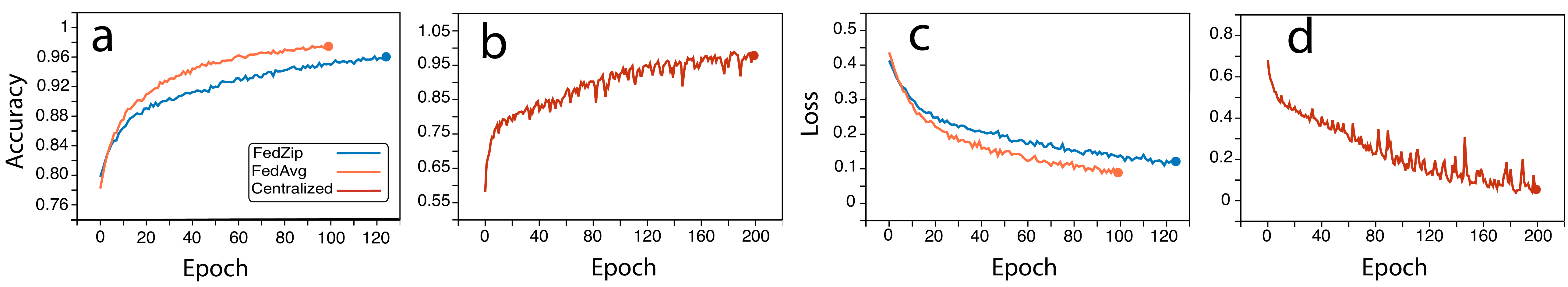}
\caption{Convergence speed for the Insight4wear dataset. The accuracy and loss for decentralized (a, c) and Centralized (b, d) approach is depicted. The FedZip with compression (blue line) and FedAvg without compression (orange line) is illustrated. With compression, the convergence speed is decreased but the accuracy reaches the baseline. The centralized approach converges after 200 local epoch, which is approximately two times more than FedAvg and FedZip.}
\label{fig:insigt}
\end{figure*}

The convergence of FedZip is decreased by around 25 rounds compared to FedAvg to reach the same training accuracy (Figure \ref{fig:insigt}). The Figures \ref{fig:insigt} a, b are the accuracy and the Figures \ref{fig:insigt} c, d indicates the loss. the The degradation in convergence and test accuracy is caused by insufficient number of data points and thus results in overfitting. By augmenting the dataset and adding noise, we may have achieved the same result with the same convergence speed, but reporting with augmentation makes the comparison biased.

\begin{table}[htbp]
\begin{center}

\caption{Compression in different decentralized approaches. We obtain a $1085\times$ (the highest) compression rate near 1 percent degradation in test accuracy. } \label{tab:cmpDiffApp}
\begin{tabular}{|c|c|c|c|}
\hline
\textbf{Methods} & \textbf{Compression} & \textbf{Test Accuracy} & \textbf{Convergence} \\
  & \textbf{Rate}& \textbf{(Degradation)\footnotemark} & \textbf{Speed}(Round) \\
  \hline
\midrule
    \ FedAvg &	1x & baseline\% & 20\\
    \hline
    \ TernGrad & 16x & -0.92\%  & 25\\
    \hline
    \ SignSGD & 32x & -1.5\% & 25\\
    \hline
    \ GD & 50x & -1.5\% & 20\\
    \hline
    \ DGC &  597x & -1\% & 25\\
    \hline
    \ STC & 1050x &  \textbf{-0.67\%} & 25\\
    \hline
    \ \textbf{FedZip} & \textbf{1085}x & -1.12\% & 20 \\
    \hline
 \bottomrule
\end{tabular}
\end{center}
\end{table}

\begin{table}[htbp]
  \centering
  \setlength\tabcolsep{2pt}
  \label{tab:freq}
\caption{Baseline approaches are compared with FedZip on a real world dataset, Insight4Wear dataset, for 50 clients. Note that FedZip does not reduce the number of parameters; it encodes $\Delta w$ with a lossless approach to reduce the size of communication updates. }

\begin{tabular}{|c|c|c|c|c|c|c|}
\hline
\textbf{Methods} & \textbf{Convergence} & \textbf{Training}& \textbf{Loss} & \textbf{Test} & \textbf{CR} & \textbf{Size (MB)} \\
  & \textbf{Round} & \textbf{Accuracy}& & \textbf{Accuracy} &  & \\
  \hline
    \ Centralized & 200 (epochs) & 97.82 & 0.0543 & 86.77 & 1x & 0.7192\\
    \hline
    \ FedAvg &	100 &	97.46 &	0.0890 & 77.69 & 1x & 0.7192\\
    \hline
    \ FedSgd & 250& 96.49 & 0.1261 & 76.38& 1x & 0.7192 \\
    \hline
    \ \textbf{FedZip} & 125  &	96.53 &	0.1213 & 75.93 & avg 145x & 0.0051\\
 \hline
  \end{tabular}
  
  \label{tab:insight4wear}
\end{table}

\subsection{Compression}
The ultimate objective of our work is to reach the highest compression rate possible while maintaining convergence and accuracy. To quantify this objective, we need to record the number of $\Delta w$ sizes sent from clients to the server. Because of the randomness of the number of weights in each cluster, the compression rate varies. Therefore, we report a range of recorded $\Delta w$ sizes (update sizes) in different rounds and for random clients in Figure \ref{fig:histrec}. 

We aggregate 200 records for the CNN and 100 records for the VGG16 network. The histograms in Figures \ref{fig:myplotHuffma1}, \ref{fig:myplotHuffma2} describe the results of Huffman Encoding for CNN and VGG16, which achieves up to approximately 32$\times$ CR. The first proposed encoding stores the position of two infrequent clusters in $\Delta w$ in address tables, and attains up to a $223\times$ CR, with an average CR of $80\times$ for CNN, and a maximum of $27\times$ and average of $16\times$ for VGG16. These results can be seen in the histograms in Figures \ref{fig:myplotcnn1} and \ref{fig:myplotvgg1}. By storing differences of position from two infrequent clusters in address tables, we accomplished a maximum of $1085\times$ CR and an average of $612\times$ CR for CNN, and a maximum of $194\times$ CR and average of $127\times$ for VGG16. These results are illustrated in Figures \ref{fig:myplotcnn2} and \ref{fig:myplotvgg2}. In a more technical sense, the 4.7 MB size of $\Delta w$ in CNN is compressed to 0.0078 MB on average, with the best case 0.0044 MB for 200 records (Figure \ref{fig:histrec}).

Table \ref{tab:cmpDiffApp} shows the Test Accuracy, Compression Rate, and Convergence Speed. Table \ref{tab:cmpDiffApp} reports the results of the information between FedZip and other state-of-the-art methods. FedZip significantly outperforms the state-of-the-art compression rate by compressing $b_{total}$ without including other elements from Equation \ref{eq:compression}, such as \textit{E} and \textit{N}.\par 

\begin{figure*}[t]
  \centering
  \captionsetup[subfigure]{labelformat=empty}
  \subcaptionbox{{\fontfamily{phv}\selectfont
    \large{a}}\label{fig:T_CNN}}[.24\linewidth][c]{%
    \includegraphics[width=\linewidth]{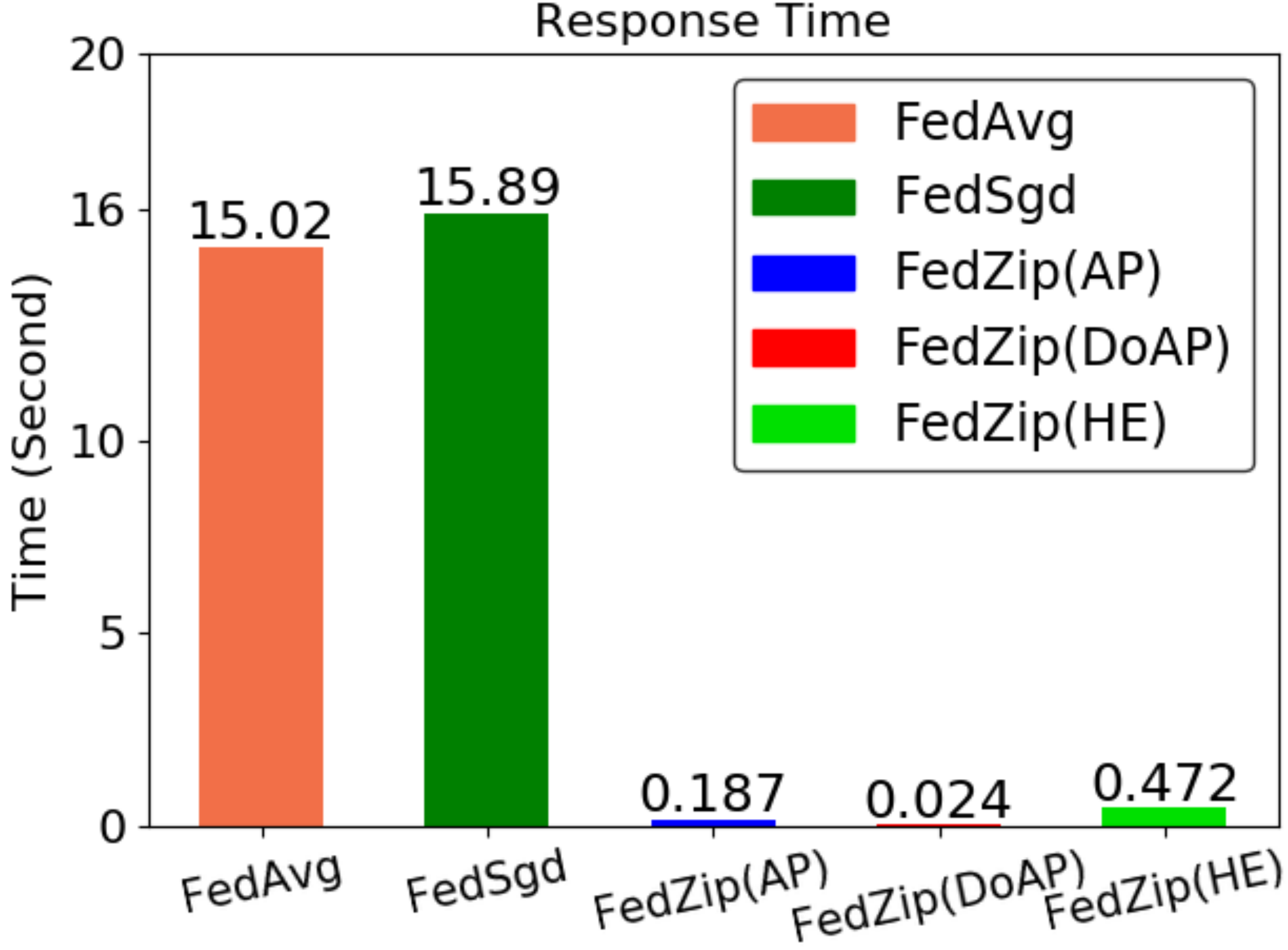}}
  \subcaptionbox{{\fontfamily{phv}\selectfont
    \large{b}}\label{fig:T_VGG}}[.24\linewidth][c]{%
    \includegraphics[width=\linewidth]{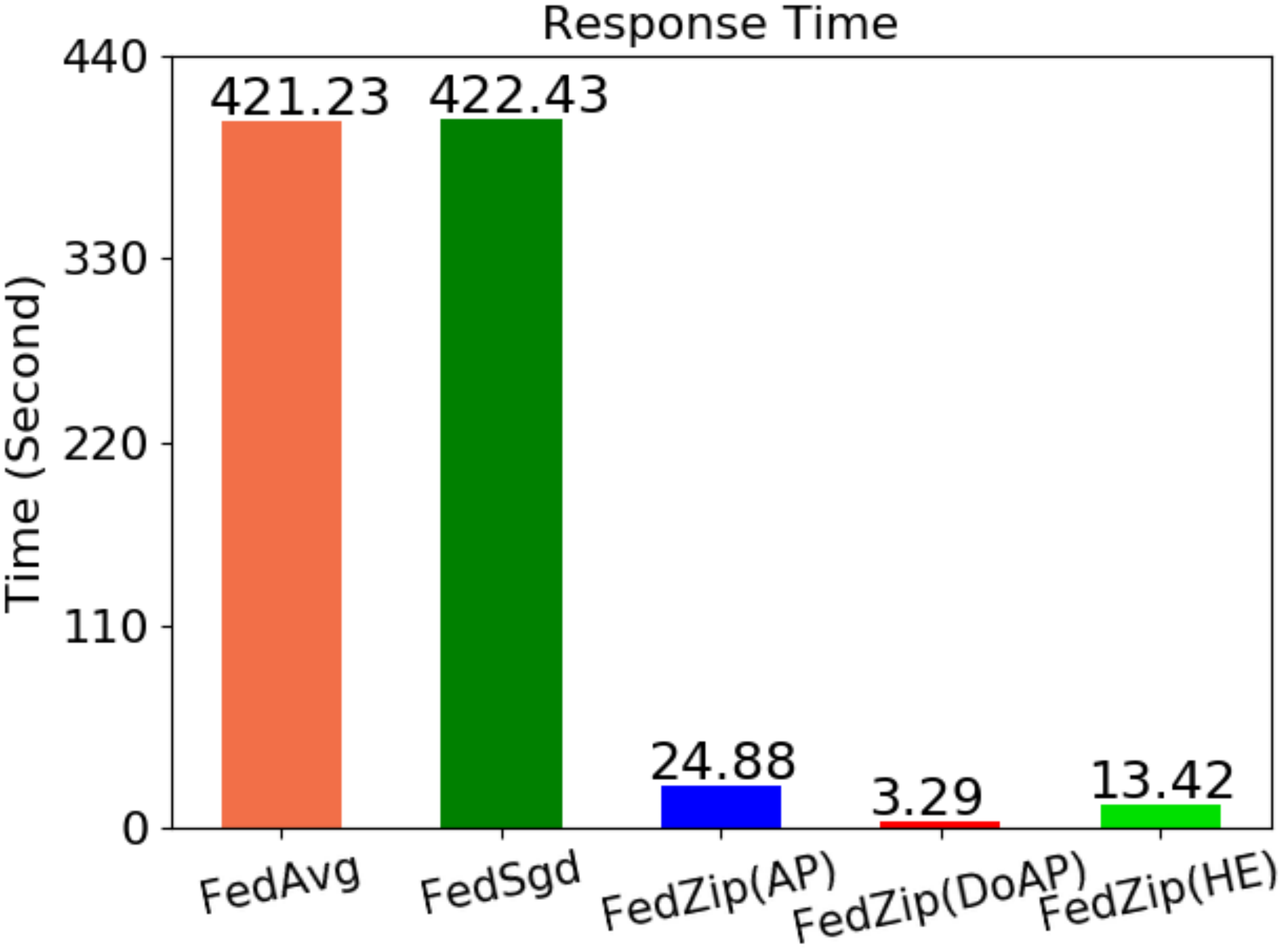}}
  \subcaptionbox{{\fontfamily{phv}\selectfont
    \large{c}}\label{fig:B_CNN}}[.24\linewidth][c]{%
    \includegraphics[width=\linewidth]{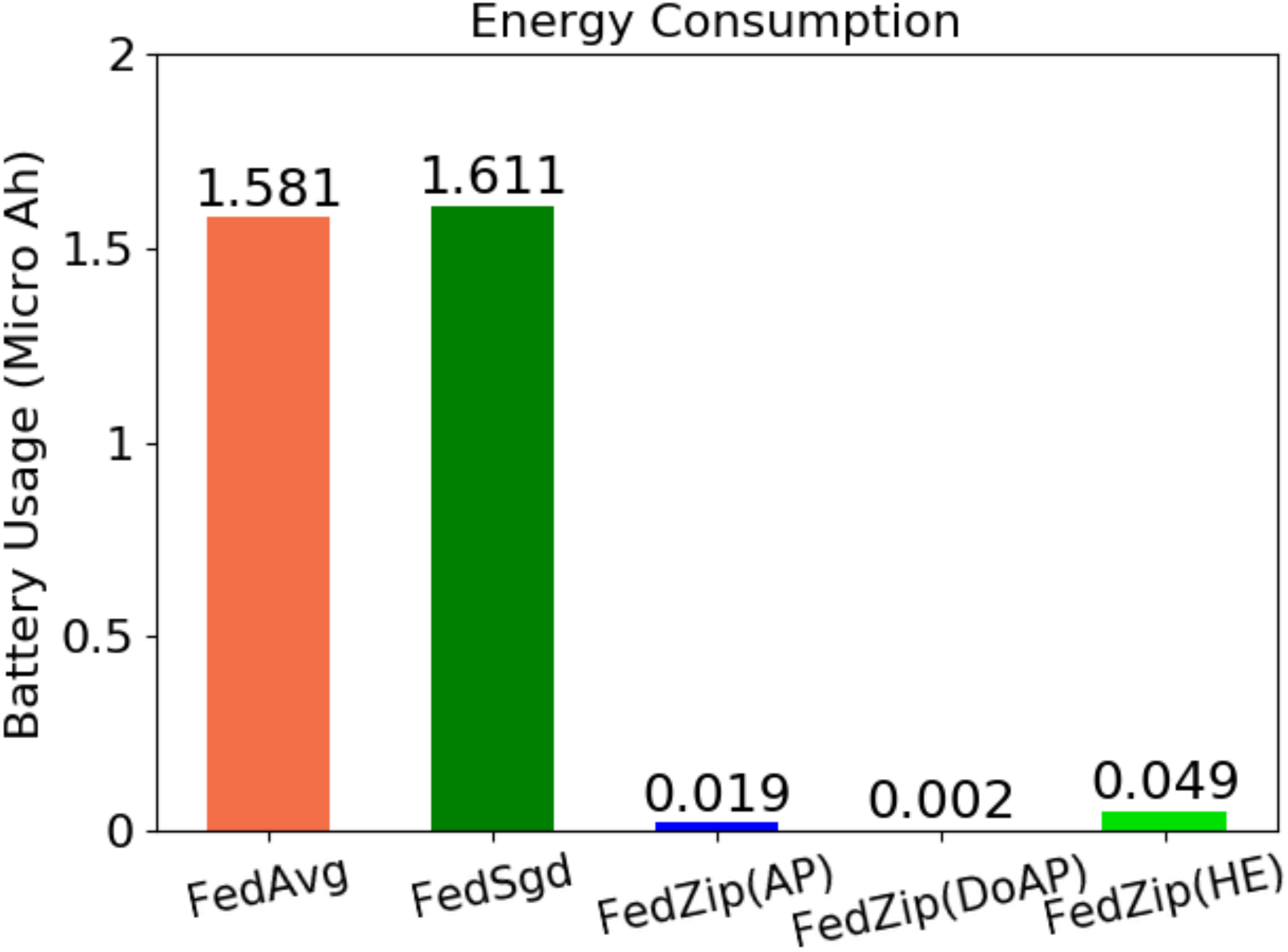}}
  \subcaptionbox{{\fontfamily{phv}\selectfont
    \large{d}}\label{fig:B_VGG}}[.24\linewidth][c]{%
    \includegraphics[width=\linewidth]{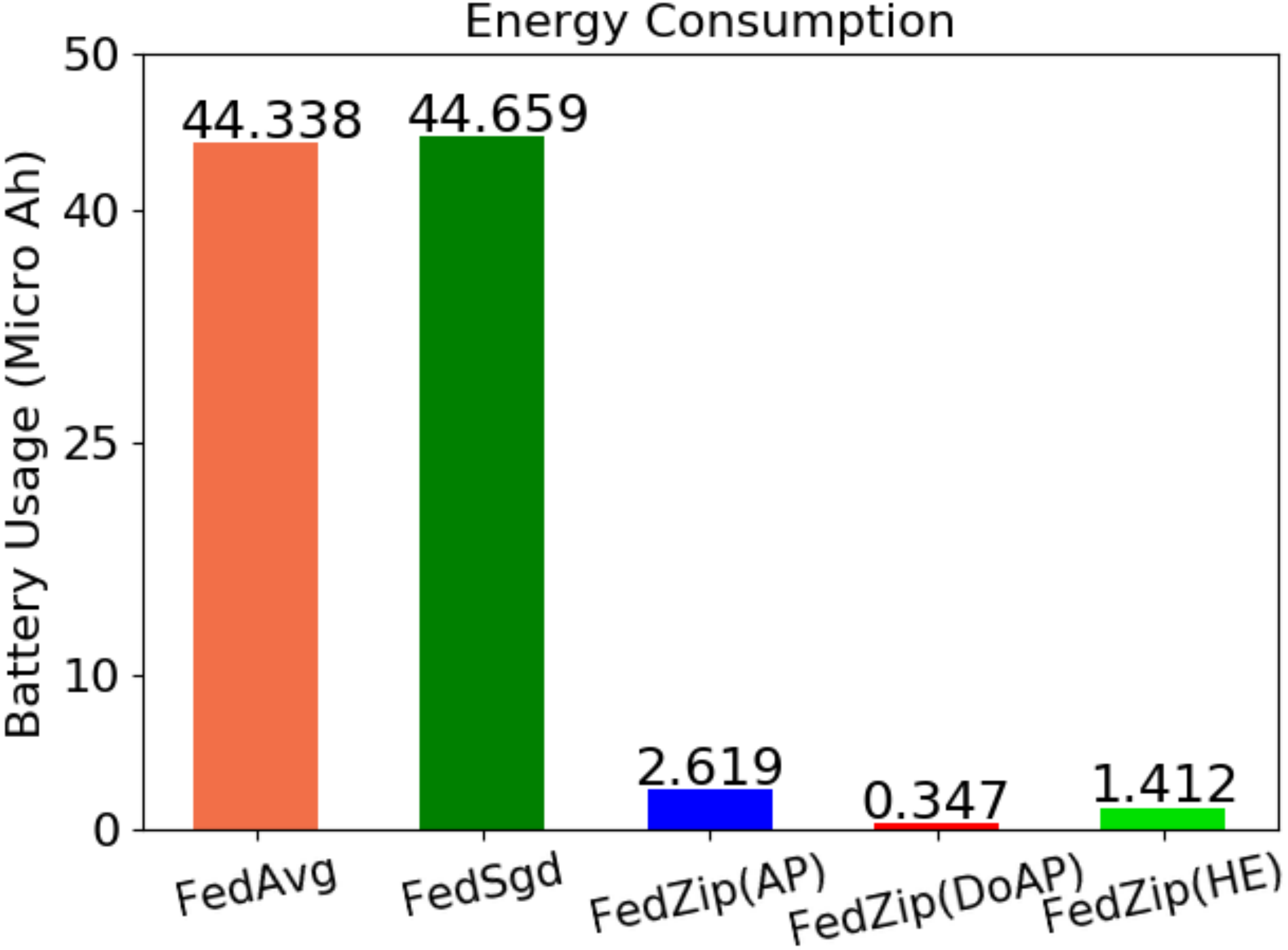}}
\captionsetup{skip=-2pt}
\caption{Resource utilization of FedZip compared with FedAvg and FedSGD. AP stands for Address Position, DoAP stands for Difference of Address Position, and HE stands for Huffman Encoding. (a) execution time (second) for CNN model, (b) execution time for VGG16 model, (c)  battery utilization in micro-Amperehour for CNN model, (d) battery utilization in micro-Ampere hour for VGG16 model.}
\label{fig:resource}
\end{figure*}
\vspace{-3pt}
\subsection{Scalability}
One of the main requirements of machine learning algorithms designed for mobile and other battery-powered devices is scalability \cite{Rawassizadeh16}. A lack of scalability negatively affects the quality of the communication they are performing and results in missing data \cite{reza2019b}. This problem grows more pronounced when dealing with deep learning models with millions of parameters. 

\footnotetext{The values in this column are collected from training a CNN model on MNIST data set of the mentioned models and reported as test accuracy. Note that FedAvg is the assumed baseline, and thus degradation of accuracy is the difference between each model and FedAvg.}

In this section, we analyze the scalability of three encoding methods for our framework and demonstrate the feasibility of the best case. We choose to conduct these experiments on the VGG16 neural network. It is a well-known neural network and an important test case for every compression-based improvement of the FL framework because it has so many  parameters---approximately $138.5$ million \cite{Karen14}. 

The FedZip framework uses Huffman Encoding, and our experiments demonstrate that Huffman Encoding is highly flexible to scalability, and thus can provide an approximately $~32\times$ CR \cite{Karen14}. Nevertheless, the first encoding method does not offer the highest possible CR. The CR of the second encoding, which stores the address positions of two infrequent clusters inside an address table, is decreased by less than $~27\times$. This can be seen in Figure \ref{fig:resource} as well. This large decrease is due to the storage of the exact position in an address table and demonstrates that our second encoding method is highly sensitive to scalability. However, it can be used in small networks. The third encoding method, which stores the difference of address positions in an address table, performs best in both models (CNN and VGG16). Although it shows a degradation in CR, the third method suggests a promising CR of up to $~194\times$, with an average of $~127\times$. This encoding method can be considered a sensitive approach in terms of scalability, but with much more resistance than the two other methods tested. Figure \label{fig::histrec} offers a comparison of scalability between the three proposed encoding methods. Figure \ref{fig:myplotvgg1}, \ref{fig:myplotvgg2} presents a histogram of $100$ update records with $\Delta w$ sizes for the VGG16 model \cite{Karen14}. The CR is multiplied by a factor $E$ (the training step in each client). For example, with $E=10$, FedZip framework can achieve up to a $10850\times$ CR. However, the effect of the element $E$ (see Equation \ref{eq:compression}) is complex and will require independent study, beyond the scope of this paper, in which we focus on $b_{total}$. The FedAvg with a delayed communication techniques reaches $400\times$ CR, and this technique can also be applied to FedZip. 

Our framework imposes between a $5\%$ to $10\%$ overload in terms of time complexity in each client due to the necessary calculation for sparsification, clustering, and encoding, which is cost-effective in comparison with the time saved when sending the update.\par

\vspace{-3pt}
\subsection{Resource Utilization}
A known issue among mobile and ubiquitous devices is their limited capabilities due to battery capacity \cite{reza2014}. In comparison to desktop computers, mobile phones, and other portable computers suffer from the limited power supply and  slower response time. 
Figure \ref{fig:resource} reports differences in battery utilization and response time, between  the FedAvg, FedSGD, and FedZip frameworks. Note that as the number of packet sizes increases (which maps to model parameters) the differences between the battery efficiency of FedZip and baseline (FedAvg) increase further. The results of our experiment on the smartphone are shown in Figure \ref{fig:resource}. They present that FedZip is significantly more efficient, and it also runs significantly faster than two other state-of-the-art FL architectures. FedZip reduces the communication overhead between clients and a server by $\sim 99\%$ in our experiments.
Here, we provide a comparison of FedZip to FedAvg and FedSGD implemented on a smartphone. In particular, we experiment by sending $\Delta w $ of the CNN and VGG16 model as a message from clients to a server within the FedAvg, FedSGD and FedZip architectures. This experiment is conducted in a real-world situation on the deep learning model described in this work. We conduct this experiment on a Xiaomi Mi A2 smartphone running Android OS 9.0, with a 4 MB download rate and 256--512 kB upload transmission rate. We do 10 experiments on each case to measure response time and battery usage and report the average results.

\section{CONCLUSION}
In this paper, we propose FedZip, a compression framework that overcomes communication challenges, energy consumption, and high degradation in FL frameworks. Our experiments demonstrate that the FedZip is able to improve communication of FL architecture with only insignificant effects on accuracy and convergence speed. We show that our novel pipeline of algorithms, FedZip, efficiently compresses communication updates between clients and a server, with insignificant impact on the convergence of deep learning models in a distributed architecture. In future work, we will consider the global model and examine the effects on compression data during server-client communication. We will also integrate components of our framework into on-device information retrieval methods to improve the efficiency of search algorithms on devices without an edge or cloud connection.

\bibliographystyle{ieeetr}
\bibliography{sample-base}

\end{document}